\documentclass[runningheads]{llncs}
\usepackage{graphicx}
\usepackage{amsmath}
\usepackage{array}
\usepackage{epsfig}
\usepackage{tabu}
\usepackage{wrapfig}
\usepackage{float}
\usepackage{subcaption}
\usepackage[export]{adjustbox}
\newcolumntype{M}[1]{>{\centering\arraybackslash}m{#1}}
\newcolumntype{N}{@{}m{0pt}@{}}

\begin{document}
\title{DeepHuMS: Deep Human Motion Signature for 3D Skeletal Sequences}
%
%

\author{Neeraj Battan* \and
Abbhinav Venkat* \and
Avinash Sharma}
\authorrunning{N Battan, A Venkat et al.}
%
\institute{International Institute of Information Technology, Hyderabad (IIIT-H), India.
\email{\{neeraj.battan, abbhinav.venkat\}@research.iiit.ac.in, }
\email{\{asharma\}@iiit.ac.in}
}
\maketitle              
%
\begin{abstract}
3D Human Motion Indexing and Retrieval is an interesting problem due to the rise of several data-driven applications aimed at analyzing and/or re-utilizing 3D human skeletal data, such as data-driven animation, analysis of sports bio-mechanics, human surveillance etc. Spatio-temporal articulations of humans, noisy/missing data, different speeds of the same motion etc. make it challenging and several of the existing state of the art methods use hand-craft features along with optimization based or histogram based comparison in order to perform retrieval. Further, they demonstrate it only for very small datasets and a few classes. We make a case for using a learned representation that should recognize the motion as well as enforce a discriminative ranking. To that end, we propose, a 3D human motion descriptor learned using a deep network. Our learned embedding is generalizable and applicable to real-world data - addressing the aforementioned challenges and further enables sub-motion searching in its embedding space using another network. Our model exploits the inter-class similarity using trajectory cues, and performs far superior in a self-supervised setting. State of the art results on all these fronts is shown on two large scale 3D human motion datasets - NTU RGB+D and HDM05.\footnotetext[1]{Link for Code: \url{https://github.com/neerajbattan/DeepHuMS}.}\footnotetext[2]{Link for Project Video: \url{https://bit.ly/31B1XY2}.}

\end{abstract}

\begin{figure*}[h!]
\begin{center}
\includegraphics[width=0.9\textwidth]{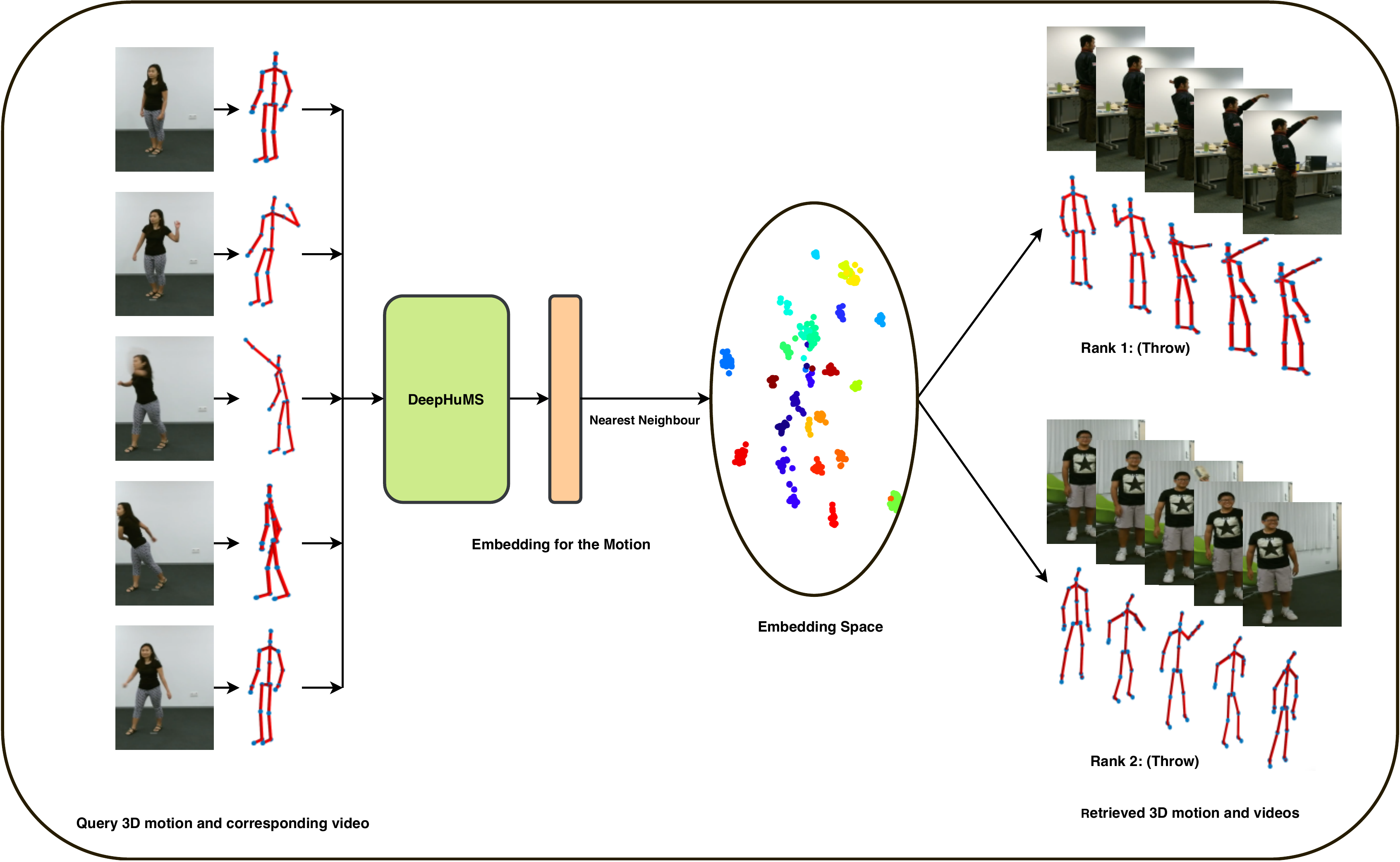}
\end{center}
\caption{Motivation of 3D Human Motion Retrieval. Given a query 3D skeletal motion sequence, we retrieve the top-k most similar sequences. A major motivation is that the corresponding videos of the retrieved results are view, appearance and background invariant.}
\label{fig:teaser}
\end{figure*}

\section{Introduction}
3D Human Motion Retrieval is an emerging field of research due to several attractive applications such as data-driven animation, athletic training, analysis of sports bio-mechanics, human surveillance and tracking etc. Performing such analysis is challenging due to the high articulations of humans (spatially and temporally), noisy/missing data, different speeds of the same action etc. Recent research in pose estimation, reconstruction~\cite{venkat2019humanmeshnet,venkat2018deep}, as well as the advancement in motion capture systems has now resulted in a large repository of human 3D data that requires processing. Moreover, since the procurement of new motion data is a time-consuming and expensive process, re-using the available data is of primary importance. To that end, we solve the problem of 3D Human Motion Retrieval and address several of the aforementioned challenges, using, a 3D human motion descriptor learned using a deep learning model.

While 3D human motion recognition is a commonly researched field, 3D human motion retrieval is much less explored. The task of human motion retrieval consists of two parts - building the feature representation and then the retrieval algorithm. Therefore, it requires recognizing the action as well as, importantly, enforcing a ranking i.e., a "low-dimensional" "recognition-robust" and "discriminative" feature embedding that is capable of fast retrieval is desirable. 

Aiming at incorporating several of these properties, several hand crafted features from skeleton sequences have been developed~\cite{xiao2013motion}. There has also been considerable research in the direction of improving the retrieval algorithm~\cite{liu2017efficient} and having better similarity metrics for comparison~\cite{chen2010learning}. For retrieval purposes, one common method is to solve an optimization problem, which is however slow and susceptible to local minimas~\cite{wang2016adaptive}. Alternatively, a few others perform a histogram/code-book matching. However, these methods are affected by noisy data, different lengths and variable frame rates of sequences, etc. Moreover, they all demonstrate their retrieval accuracy over a very small number of sequences and classes. Hence, we would like to move towards learnable representations that can account for several of these shortcomings, while still maintaining minimal supervision.

A closely related problem to retrieval in which learnable representations have been widely explored is 3D action/motion recognition. In the last few years, several deep learning model innovations have been made to better exploit the spatial and temporal information available in skeleton data ~\cite{Li:2016:ARL:2911996.2912001,Li_2018,tang2018deep}. While these models do a respectable job in recognition, they perform poorly in retrieval due to not having a discriminative enough embedding space. Further, several of them highly depend on the availability of class labels. The number of class labels available in existing datasets is fairly limited, and such supervised models are incapable of exploiting similar sub-actions amongst various classes. Hence, the requirement of a more generalized model is in order. 

Therefore, in this paper, we would like to propose a discriminative learnable representation, DeepHuMS, for retrieval, which produces instantaneous retrieval with a simple nearest neighbour search in the repository. To summarize, our contributions are:
\begin{itemize}
    \item We propose a novel deep learning model that makes use of trajectory cues, and optionally class labels, in order to build a discriminative and robust 3D human motion descriptor for retrieval.   
    \item Further, we perform sub-motion search by learning a mapping from sub-sequences to longer sequences in the dataset by means of another network.
    \item Experiments are performed, both, with and without class label supervision. We demonstrate our model's ability to exploit the inter-class motion similarity better in the unsupervised setting, thus, resulting in a more generalized solution. 
    \item Our model is learned on noisy/missing data as well as motions of different speeds and its robustness in such scenarios indicates its applicability to real world data.     
    \item A comparison of our retrieval performance with the publicly available state of the art in 3D motion recognition as well as 3D motion retrieval on 2 large scale publicly available datasets is done to demonstrate the state-of-the-art results of the proposed model. 
\end{itemize}

\section{Related Work}
Most approaches on the 3D human motion retrieval have focused on developing hand crafted features to represent the skeleton sequences\cite{xiao2013motion,wang2016adaptive,wang2013learning}. In this section, we broadly categorize them by the method in which they engineer their descriptors.
Some existing methods use objective function\cite{wang2016adaptive}, few others use codebook or histogram comparisons~\cite{liu2017efficient,kapsouras2014action} to obtain hand-crafted features. The traditional frame based approaches extract out features for every frame. 
\cite{chen2010learning} proposed a geometric pose feature to encode pose similarity. \cite{kapsouras2014action} used joints' orientation angles and angles-forward differences as local features to create a codebook and generate a Bag of Visual Words to represent the action. 
\cite{choi2012retrieval,chao2011human} suggest hand drawn sketch based skeleton sequence retrieval methods.
On the other end of the spectrum, sequence based motion features utilize global properties of the motion sequence  \cite{junejo2010view,ramezani2018motion,muller2006motion}. Muller et. al. \cite{muller2006motion} presented the motion template (MT) in which motions of the same class can be represented by an explicit interpretable matrix using a set of boolean geometric feature. To tolerate the temporal variance in the training process, dynamic time warping (DTW) was employed in their work. 
\cite{xia2012view} created a temporal motion model using HMM and a histogram of 3D joints descriptors after creating the dictionary. \cite{Qi:2014:RMD:2767045.2767056} applied a Gaussian Mixture Model to represent character poses, wherein the motion sequence is encoded, then they used DTW and a string matching to find similarities between two videos. 
 Recently many graph based models have been proposed to exploit the geometric structure of the skeleton data \cite{li2019actional,shi2019skeleton,stgcn2018aaai}. The spatial features are represented by the edges connecting the body joints and temporal features are represented by the edges connecting the same body joint in adjacent frames.

\begin{figure}[h!]
\includegraphics[width=\textwidth]{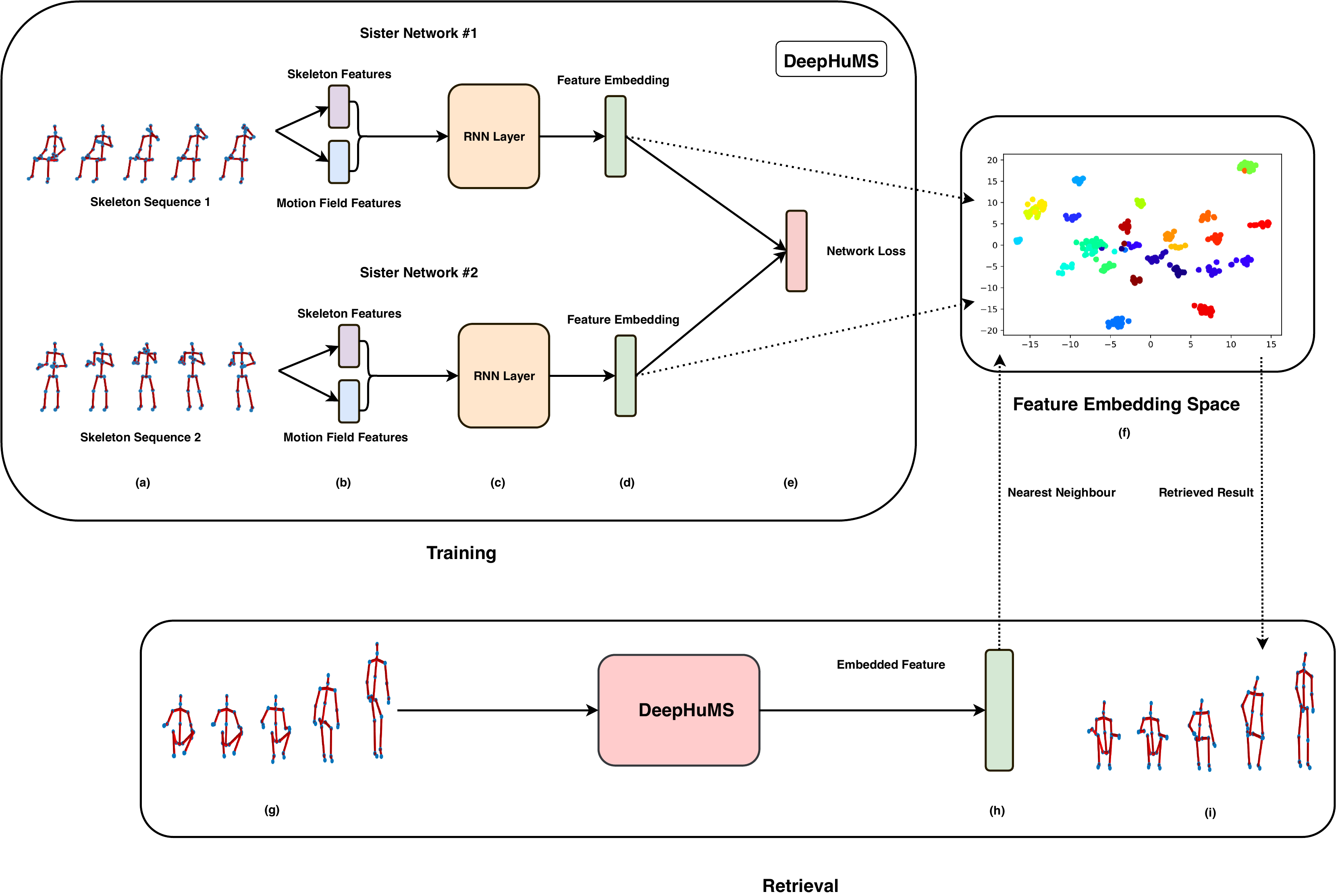}
\caption{Overview of our model - DeepHuMS. Given two skeleton sequences (a), we first extract the 3D joint locations (i.e., skeletal features) and motion field between consecutive frames (i.e., motion field features) to represent the spatio-temporal data (b). The two are concatenated together and given to an RNN~\cite{Li_2018} to model the 4D data(c). The resulting embeddings (d) are compared based using (e) contrastive loss (and optionally classification loss) to make them "discriminative" and "recognition-robust". Similarity is enforced based on the full sequence's motion distance and motion field.  At the time of retrieval, given a 3D sequence to the network (g), with the resultant embedding, a nearest neighbour search is done in the embedding space (f) generated from the training data. } 
\label{fig1}
\end{figure}

%
For the task of retrieval, \cite{wang2015deep} proposed a simple auto-encoder that captures high-level features. However, their model doesn't explicitly use a temporal construct for motion data. Primarily, learnable representations from 3D motion data have been used for other tasks. \cite{Li_2018,CLi_2018} are a few amongst many who used deep learning models for 3D motion recognition. Similarly, ~\cite{carrara2019lstm} adopts a unidirectional LSTM to encode the skeleton frames within the hidden network states and learn what subsequences of encoded frames belong to the specified action classes. 

Broadly, the existing methods are affected by noisy data, the length and variable frame rates of sequences, and are slow at retrieval. Further, they lack a learned discriminative embedding which is capable of performing sub-sequence retrieval.

\section{DeepHuMS: Our Method}
In order to build a 3D human motion descriptor, we need to exploit the spatio-temporal features in the skeletal motion data. Briefly, we have three key components - (i) the input skeletal location and joint level motion trajectories to the next frame, (ii) an RNN to model this temporal data and (iii) a novel trajectory based similarity metric (explained below) to project similar content together using a Siamese architecture. We use two setups to train our model - (a) self-supervised, with a "contrastive loss" given by Equation~\ref{eqn:contrastive} to train our Siamese model and (b) supervised setup, with a cross entropy on our embedding, in addition to the self-supervision. Refer to Figure~\ref{fig1} for a detailed architecture explanation.
\begin{equation}
    L_{contrastive} = (1-Y) \frac{1}{2} (D_{w}^{2}) + (Y) \frac{1}{2} \{max(0, m - D_{w}\}^{2}
    \label{eqn:contrastive}
\end{equation}

In Eq.~\ref{eqn:contrastive},  $Dw$ is the distance function (e.g., "Euclidean distance"), $m$ is the margin for similar and dissimilar samples and $Y$ is if the label value (1 for similar samples and 0 for dissimilar).

\begin{equation}
~\label{eqn:class}
    L_{cross entropy} = \sum_{n=1}^{M}y_{o,c}log(p_{o,c})
\end{equation}

In Eq.\ref{eqn:class}, $y$ indicates (0 or 1) if class label $c$ is the correctly classified, given $o$, the observation. $M$ is the number of classes and $p$ is the predicted probability, given an observation $o$ of class $c$.

\subsubsection{Similarity Metric.}\label{sec:losses} 
\begin{figure}[h!]
\begin{center}
\includegraphics[width=0.33\linewidth]{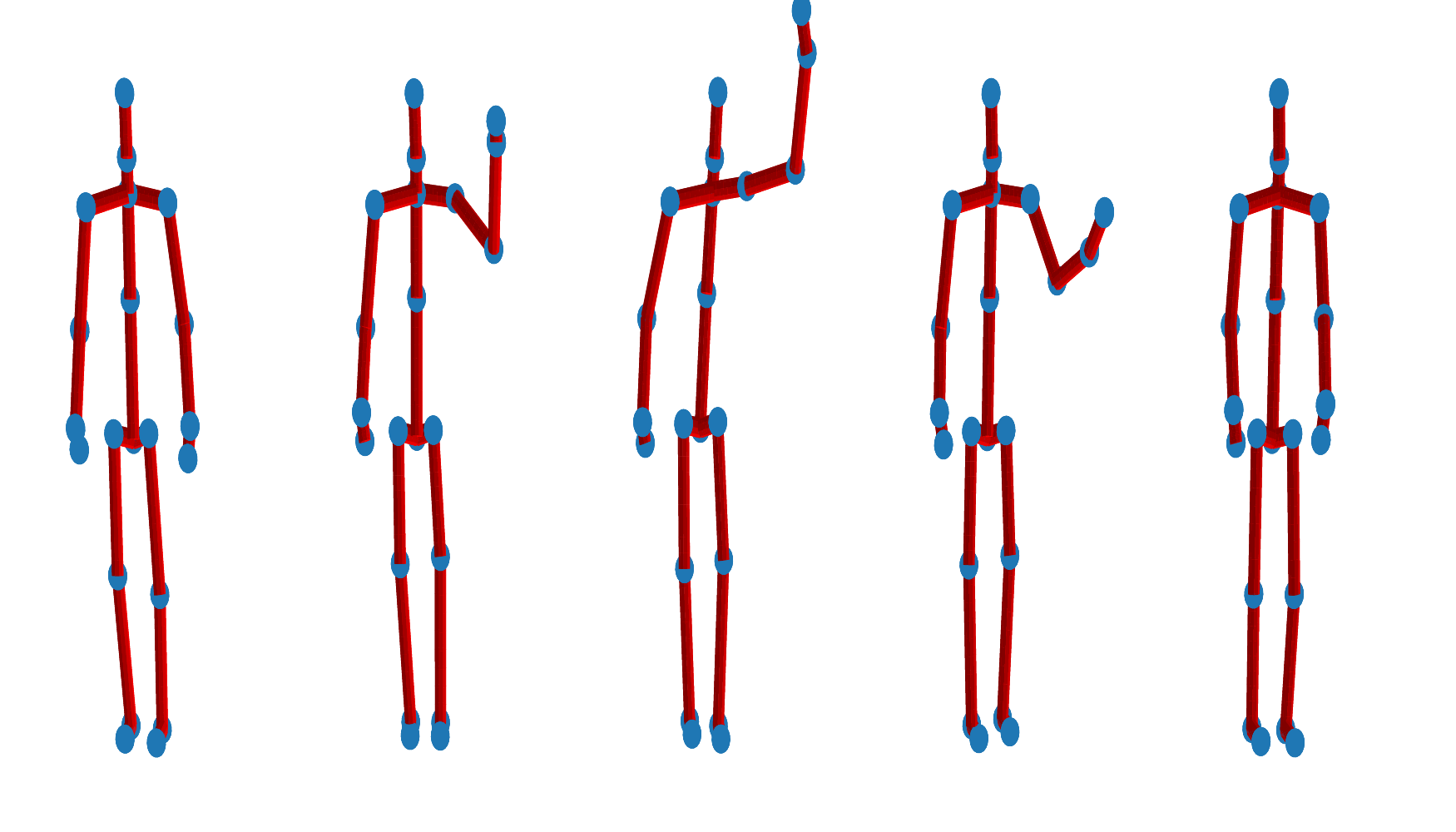}
\end{center}
\caption{Hand and elbow are important joints while waving, due to higher motion distance.}
\label{fig:waving}
\end{figure}

Two 3D human motion sequences are said to be similar if both the joint-wise "Motion Field" and joint-wise "Motion Distance" across the entire sequence are similar. The motion field depicts the direction of motion as well as the importance of the different joints for that specific sequence. The motivation behind this is evident in Figure~\ref{fig:waving} in which the hand and elbow joints are more important for $waving$. However, the motion field can end up being zero as shown in Figure~\ref{fig:waving}. Therefore, we couple it with the joint-wise motion distance in order to build a more robust similarity metric. It is to be noted that having such a full video trajectory based similarity makes it difficult to directly retrieve sub-sequences of similar content. We handle this scenario in Section~\ref{sec:discussion} using a second network. 

 \noindent  Eq.~\ref{eqn:mf} gives the motion field $MF$ between two frames $i$ and $j$. It is to be noted that we used motion field in two ways - one between every pair of frames on the input side, and the second between the first and last frame (whole video), for the similarity loss in Siamese network. Here $F[i]$ contains the 3D joints for \textit{i}th frame in the skeleton sequence. Similarly, equation~\ref{eqn:md} gives the motion distance of the entire sequence. $MD[j]$ is the total distance covered by $j^{th}$ joint and N is the number of frames in the 3D skeleton sequence.

\begin{equation}
\label{eqn:mf}
MF[i, j] = F[i]-F[j]
\end{equation}

\begin{equation}
\label{eqn:md}
    MD[j] = \sum_{i=1}^{N-1} \lVert F[i+1][j]-F[i][j] \rVert 
\end{equation}

\subsubsection{Different number of frames.} In case of sequences that have different speeds of motion or sampling rate, but similar content, the information available at the input is different, but, the resulting motion field and motion distance across the entire sequence is the same. Hence, we augment our data and enforce such sequences to be projected together in our embedding space using the contrastive loss. In other words, we map sequences with less information to the same location to sequences with more information, in the embedding space (See section~\ref{sec:discussion} for more on implementation details).

\section{Experiments}
\subsection{Datasets}

We use two commonly used large scale public MoCap datasets to evaluate our method for human 3D motion retrieval.  \\

\noindent \textbf{NTU RGB+D} ~\cite{shahroudy2016ntu}: This dataset provides RGB, depth, infra red images and 3D locations of 25 Joints on the human body. It consists of around 56,000 sequences from 60 different classes acted by 40 performers. We use the given performer wise split for learning from this dataset. \\

\noindent \textbf{HDM05} \cite{muller2007documentation}: This dataset provides RGB images and 3D locations of 31 Joints in human body. There are around 2300 3D sequences of 130 different classes performed by 5 performers in this dataset. We follow ~\cite{wang2016adaptive} for evaluation and therefore combine similar classes (for e.g. \textit{walk2StepsLstart} and \textit{walk2StepsRstart}) to get a total of 25 classes. We follow a performer-wise split with the first 4 performers for training and the last one for testing.
\subsection{Implementation Details}
All of the trained models, code and data shall be made publicly available, along with a working demo. Please refer to our supplementary video for more results.\\

\noindent \textbf{Data Pre-processing \& Augmentation.} In order to make it performer/character invariant, we normalized the 3D joint locations based on the bone length of the performer. To diversify our datasets, for every 3D sequence, we create two more sequences - a faster and a slower one. The faster sequence is created by uniformly sampling every other frame, and the slower sequence is created by interpolating between every pair of frames. 
\newline

\noindent \textbf{Network Training. }We use Nvidias GTX 1080Ti, with 11GB of VRAM to train our models. A batch size of 128 is used for NTU RGB+D dataset, and a batch size of 8 is used for training the HDM05 dataset. We use the ADAM optimizer with an initial learning rate of 10$^{-3}$, to get optimal performance on our setup. The training time for NTU RGB+D dataset is 6 hours and HDM05 is 1 hour. Each dataset is trained individually from scratch. 

\subsection{Evaluation Metrics}

\subsubsection{Retrieval Accuracy. } This is a class-specific retrieval metric. In "top-n" retrieval accuracy, we find out how many of the "n" retrieved results belong to the same class as the query motion.

\subsubsection{Dynamic Time Warping (DTW) distance. } Inspired from ~\cite{chen2010learning}, we use Dynamic Time Warping as a quantitative metric to find out the similarity between two sequences based on distance. Two actions with different labels can be very similar, for example, drinking and eating. Likewise, the same class of actions performed by two actors can have very different motion. Hence using only the class-wise retrieval accuracy as metric doesn't provide the complete picture, and therefore, we use DTW as well.

\subsection{Comparison with State of the Art}~\label{sec:comparison}Since all of the existing state of the art methods use supervision, we compare our supervised setup with them in two ways - (a) with existing 3D Human Motion Retrieval models and (b) with 3D Human Motion Recognition embeddings. Class-wise retrieval accuracy of the top-1 and top-10 results are reported for the same. 

\subsubsection{3D Human Motion Retrieval.} 
\begin{figure}[h!]
    \hfil
    \begin{subfigure}[b]{0.45\textwidth}
        \includegraphics[width=\textwidth]{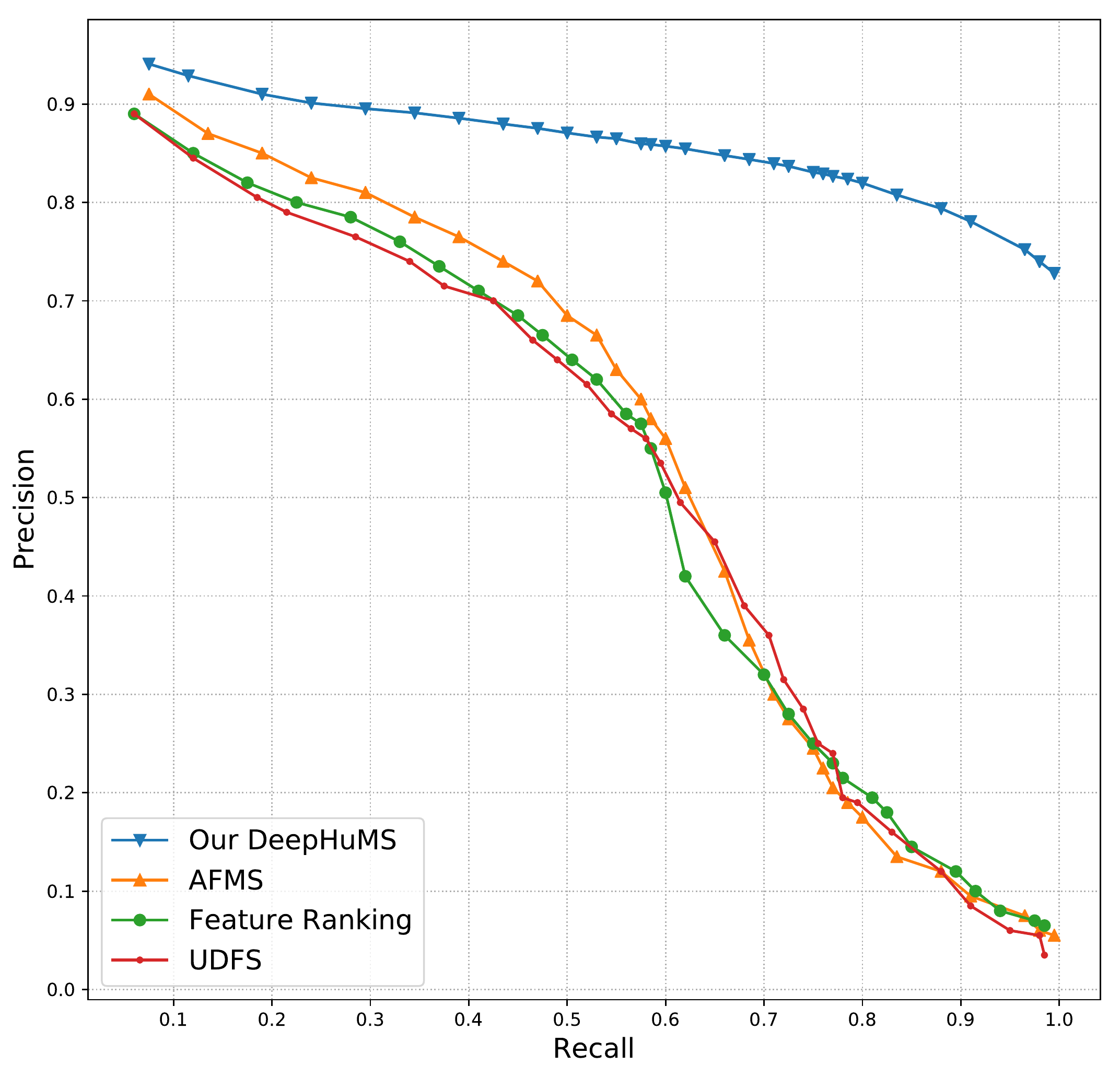}
        \caption{}
        \label{fig:retrieval}
    \end{subfigure}
    \hfil
    \begin{subfigure}[b]{0.45\textwidth}
        \includegraphics[width=\textwidth]{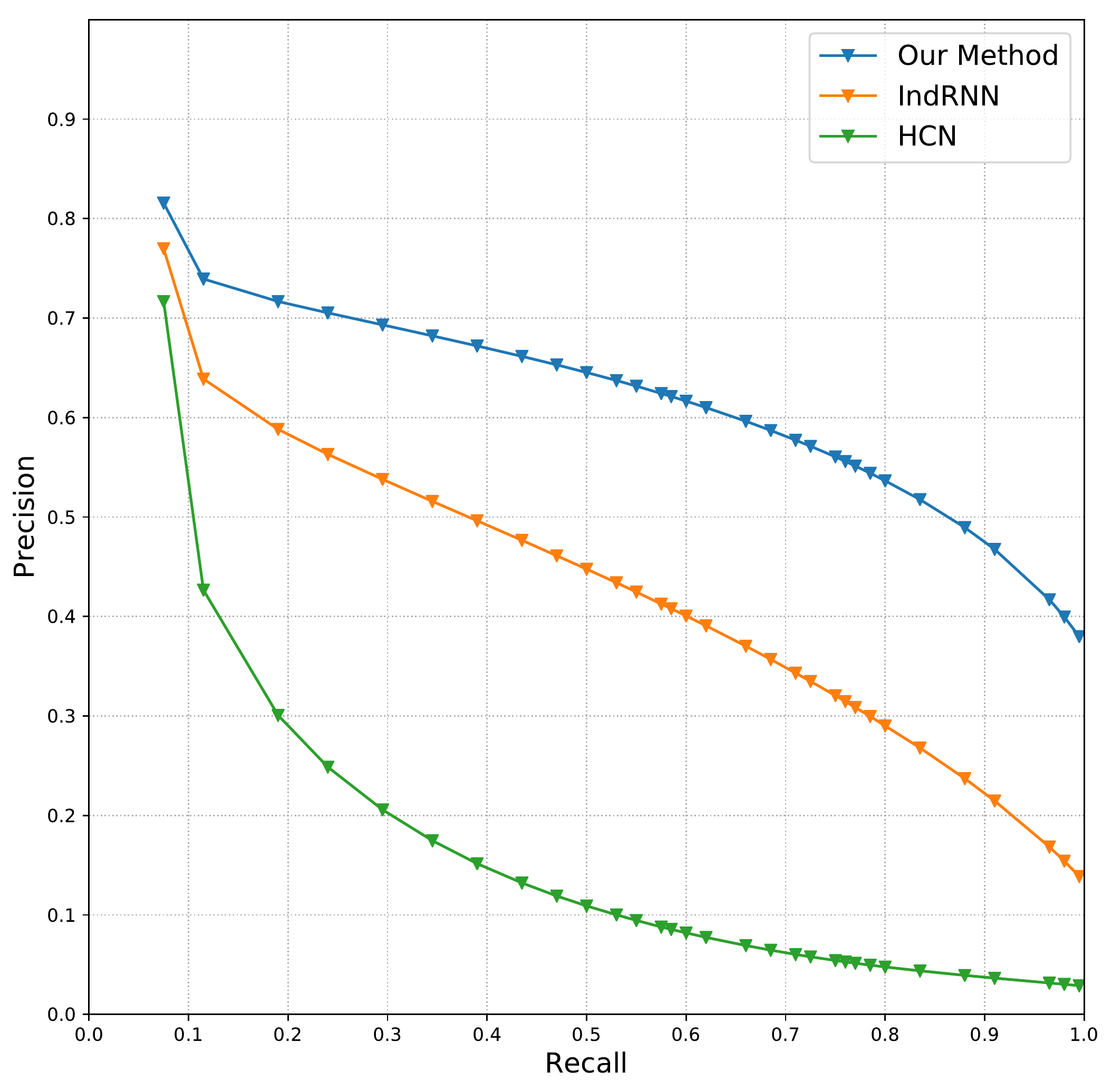}
        \caption{}
        \label{fig:pre_rec_ind}
    \end{subfigure}
    \hfil
\caption{A comparison of retrieval accuracy using PR curves for (a) 3D Motion Retrieval on HDM05 and (b) 3D Motion Recognition on NTU RGB+D}
\label{fig:tsne}
\end{figure}

Most of the existing retrieval methods~\cite{ramezani2018motion,ofli2014sequence,gowayyed2013histogram} show results on only up to 10 classes, and on very small datasets. ~\cite{wang2016adaptive} uses the same number of class labels as us, and we therefore, compare with them in Figure~\ref{fig:retrieval}. As shown in Figure~\ref{fig:retrieval}, the area under the PR curve is far larger for our method, and we have learned a much more robust 3D human motion descriptor.

\begin{table}[ht]
\begin{center}
\caption{Retrieval accuracy with 3D motion recognition on NTU RGB+D}
\begin{tabular}{| c | c | c |}
    \hline
Method & Top 1 Ret. Acc. & Top 10 Ret. Acc.\\
\hline
 HCN~\cite{CLi_2018} & 0.61 & 0.56 \\ 
\hline
 IndRNN~\cite{Li_2018} & 0.69 & 0.62 \\
\hline
\textbf{DeepHuMS (Ours)} & \textbf{0.78} & \textbf{0.753} \\
\hline

\end{tabular}
\label{tab:ret_acc}
\end{center}

\end{table}

\subsubsection{3D Human Motion Recognition. }We compare with learned representations from 3D Motion recognition. The results for the recognition models in Table~\ref{tab:ret_acc} and Figure~\ref{fig:pre_rec_ind} are computed using their embeddings trained on our datasets.

\subsubsection{Retrieval v/s Recognition.} Figure~\ref{fig:tsne} shows how our model produces a more clustered and therefore, discriminative space, suitable for retrieval, in comparison with the embedding space of ~\cite{Li_2018}, a state of the art 3D motion recognition algorithm. Recognition algorithms only focus on learning a hyperplane that enables them to identify limited motion classes. Adding a generalized similarity metric enforces an implicit margin in the embedding space - a motion trajectory based clustering.   

\begin{figure}[h!]
    \hfil
    \begin{subfigure}[b]{0.45\textwidth}
        \includegraphics[width=\textwidth]{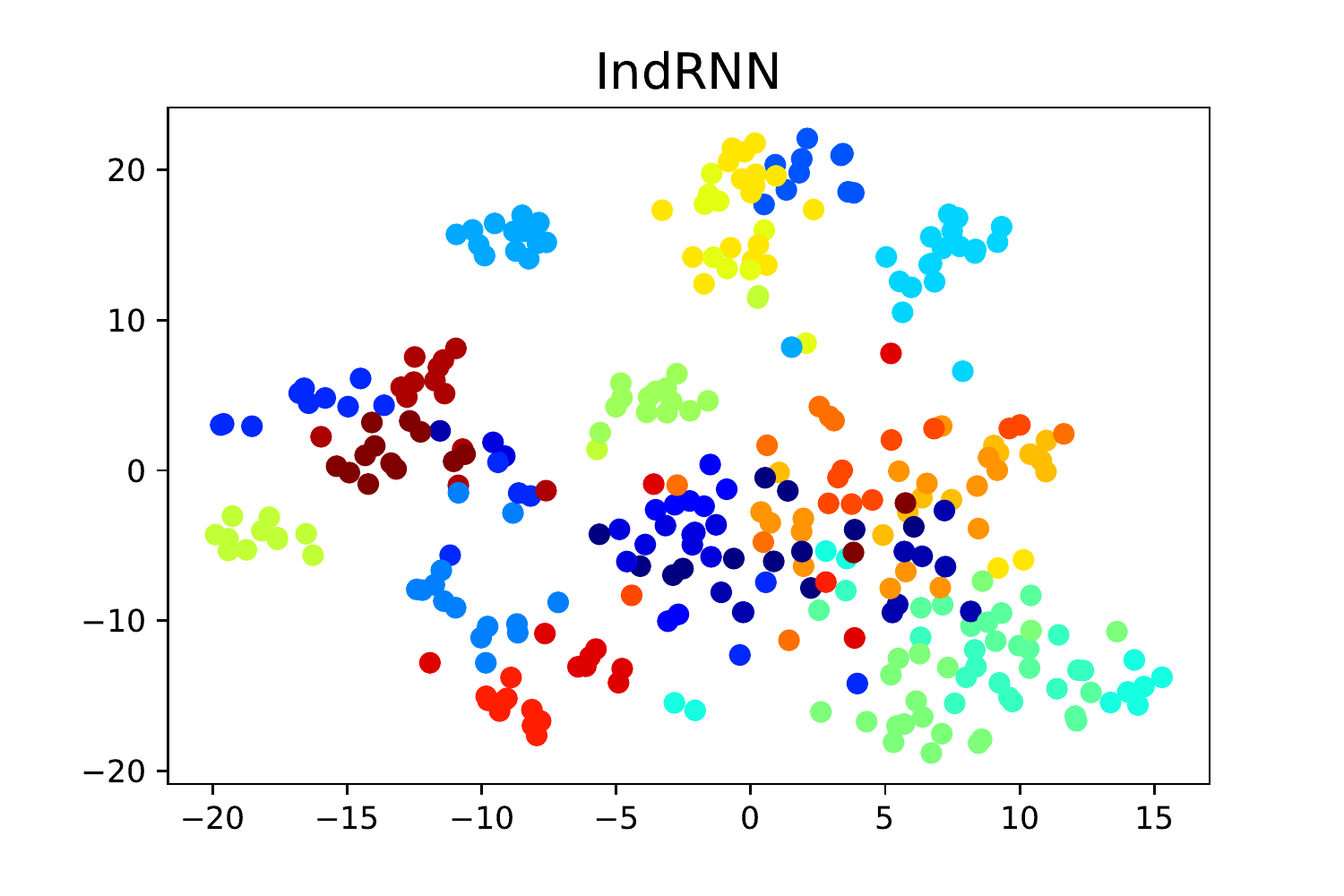}
        \caption{}
        \label{fig:tsne_or}
    \end{subfigure}
    \hfil
    \begin{subfigure}[b]{0.45\textwidth}
        \includegraphics[width=\textwidth]{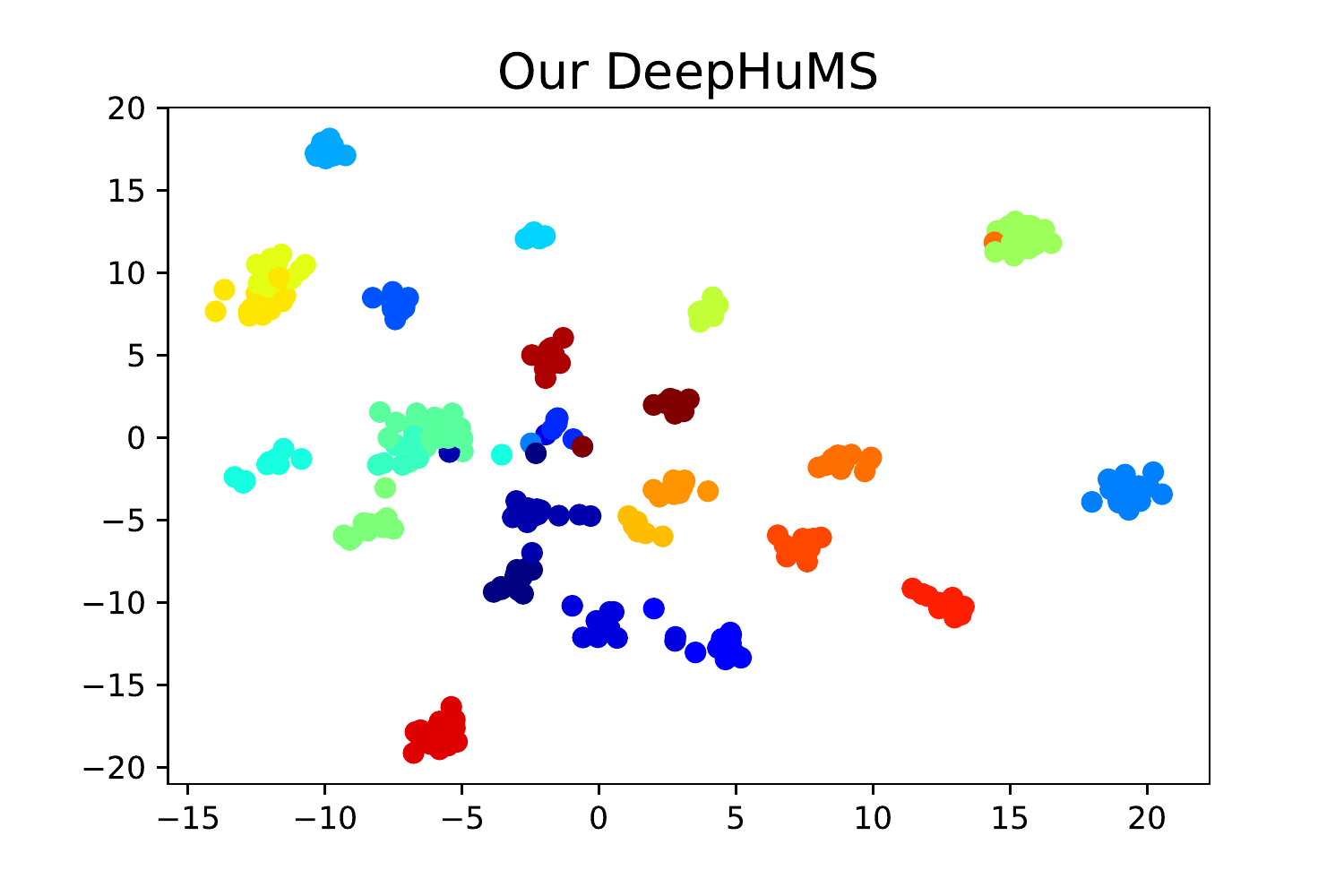}
        \caption{}
        \label{fig:tsne_up}
    \end{subfigure}
    \hfil
\caption{A comparison of the t-SNE representation of Motion Recognition \cite{Li_2018} with our method on NTU-RGB+D dataset \cite{shahroudy2016ntu}}
\label{fig:tsne}
\end{figure}

\subsection{Discussion}
\label{sec:discussion}
\begin{figure*}[h!]
\begin{center}
    \begin{subfigure}[b]{0.3\textwidth}
        \includegraphics[width=\textwidth]{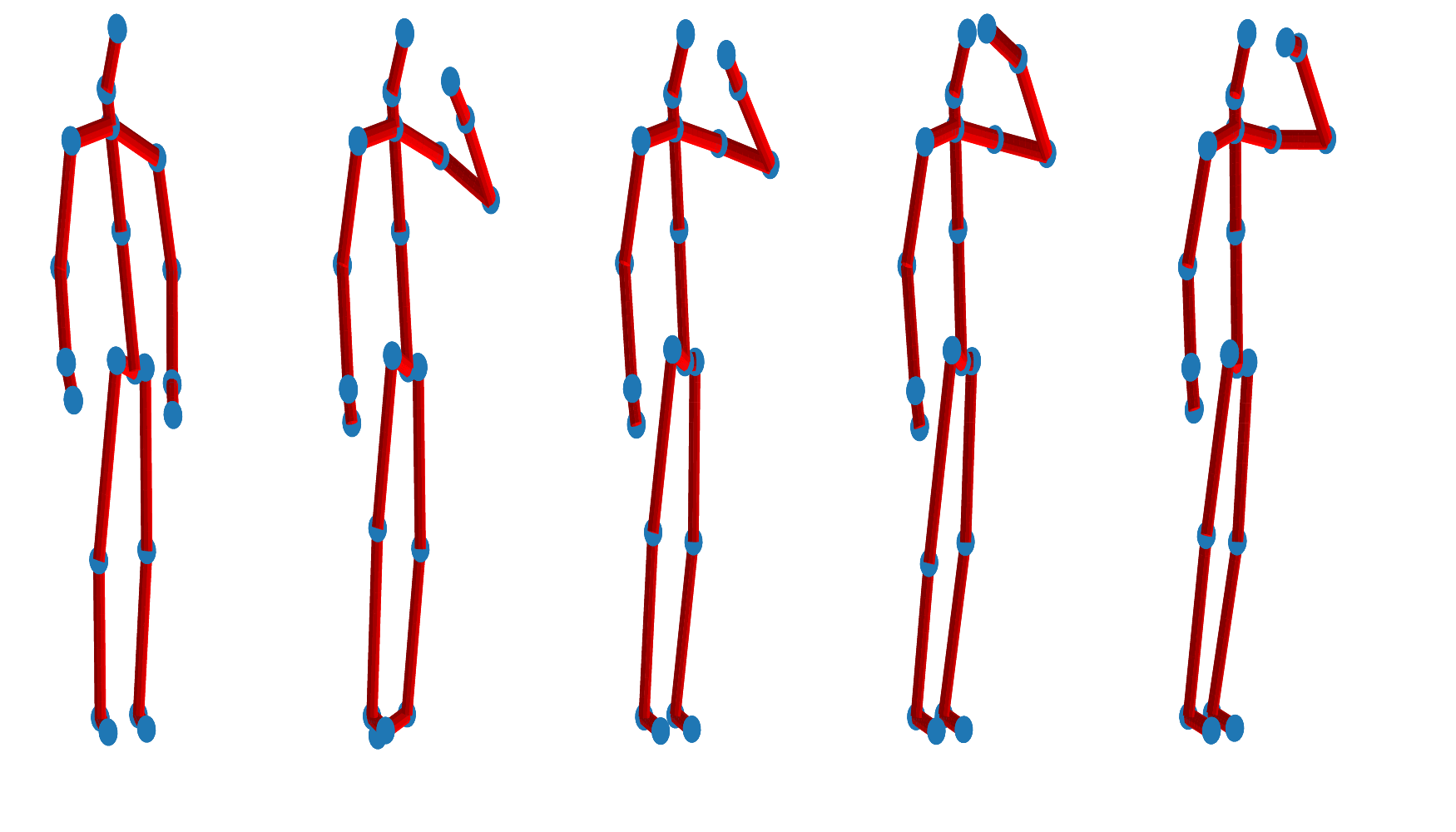}
        \caption*{Query: Drink}
        \label{}
    \end{subfigure}
\hfill
    \begin{subfigure}[b]{0.3\textwidth}
        \includegraphics[width=\textwidth]{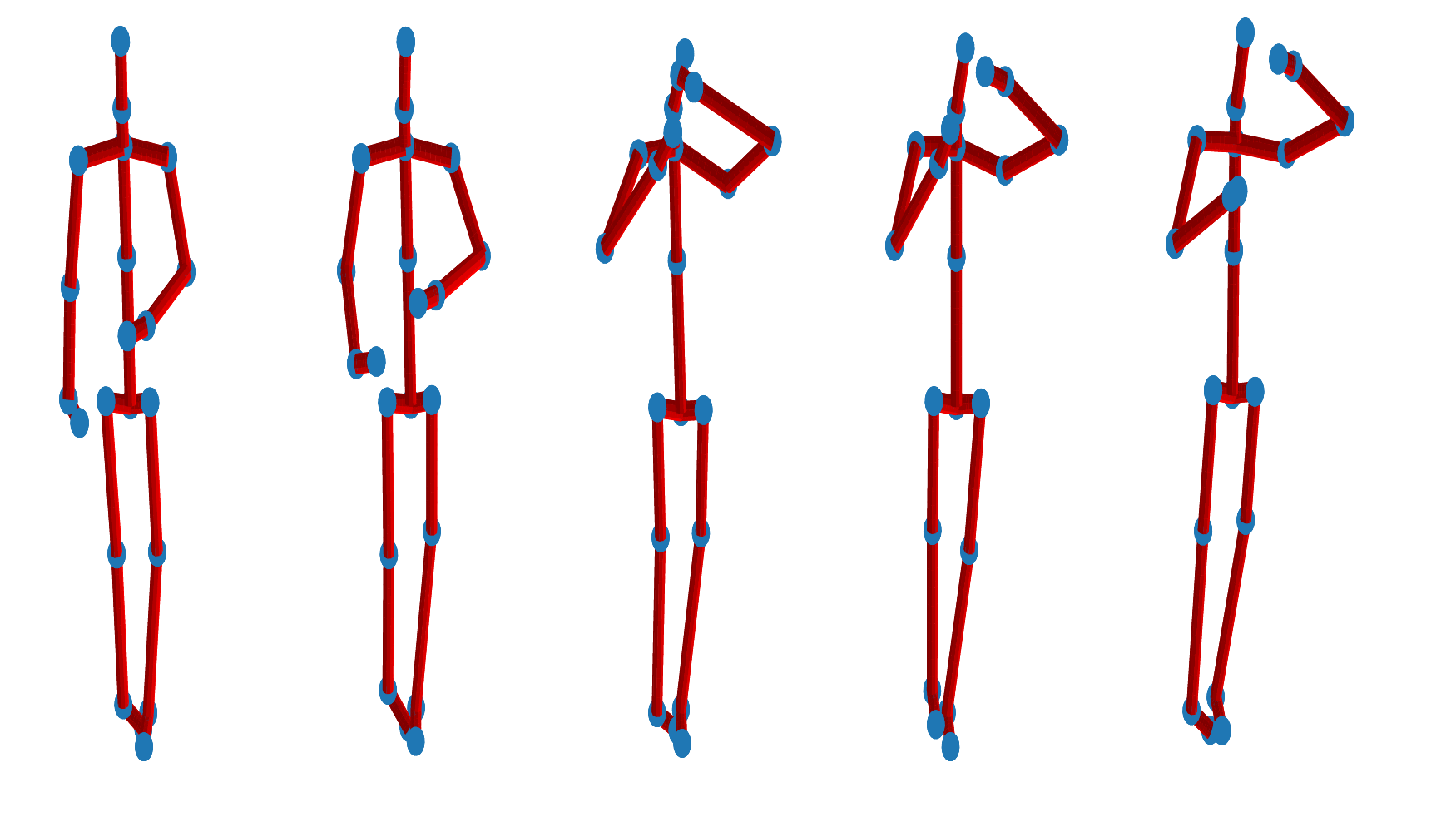}
        \caption*{Rank 1: Eat}
        \label{}
    \end{subfigure}
\hfill
    \begin{subfigure}[b]{0.3\textwidth}
        \includegraphics[width=\textwidth]{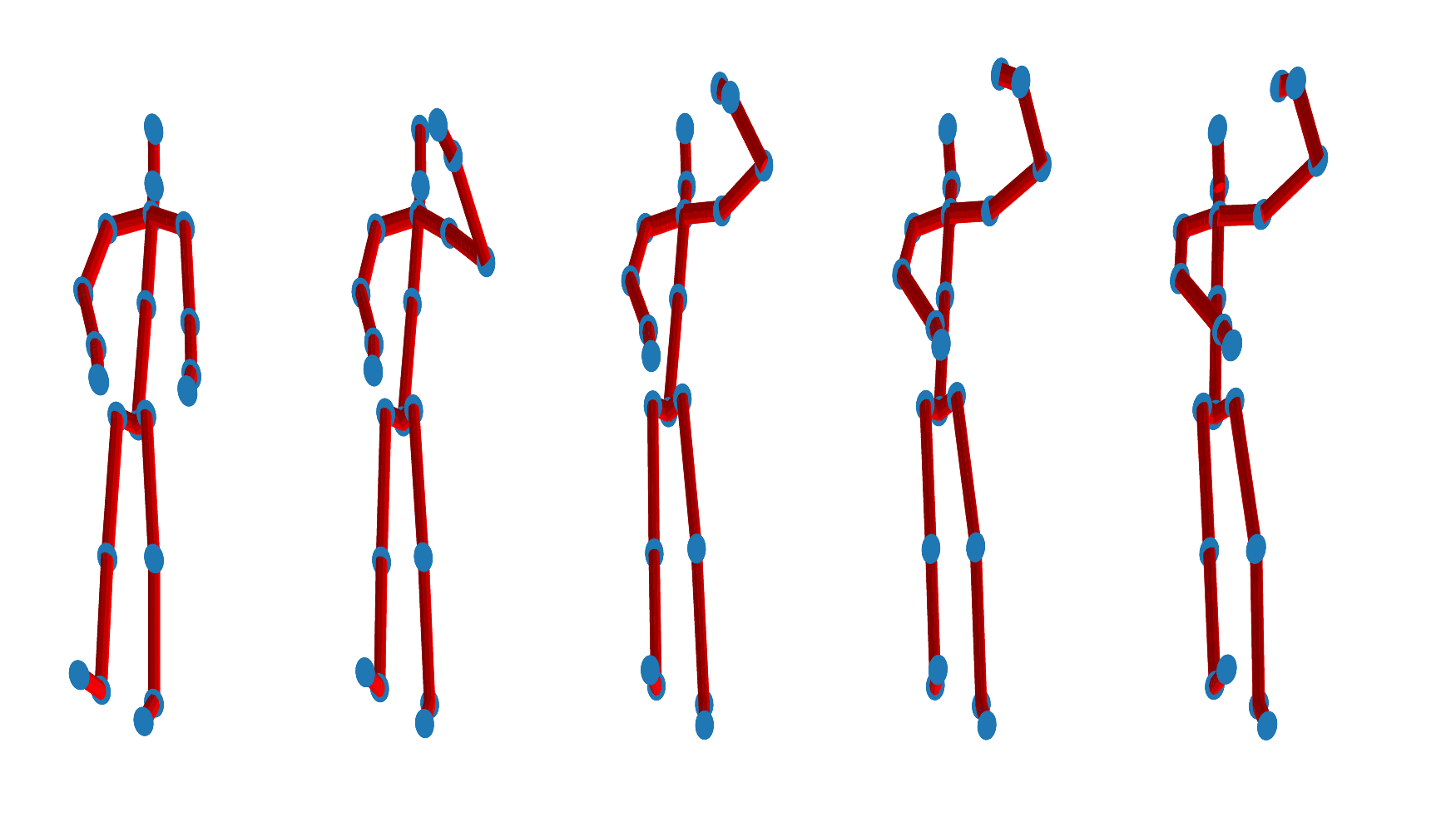}
        \caption*{Rank 2: Brush Hair}
        \label{}
    \end{subfigure}

\end{center}
\caption{Retrieval results for self-supervised setup, which shows that we exploit inter-class similarity}
\label{fig:self-sup}
\end{figure*}

\begin{figure*}[h!]
\begin{center}

    \begin{subfigure}[b]{0.3\textwidth}
        \includegraphics[width=\textwidth]{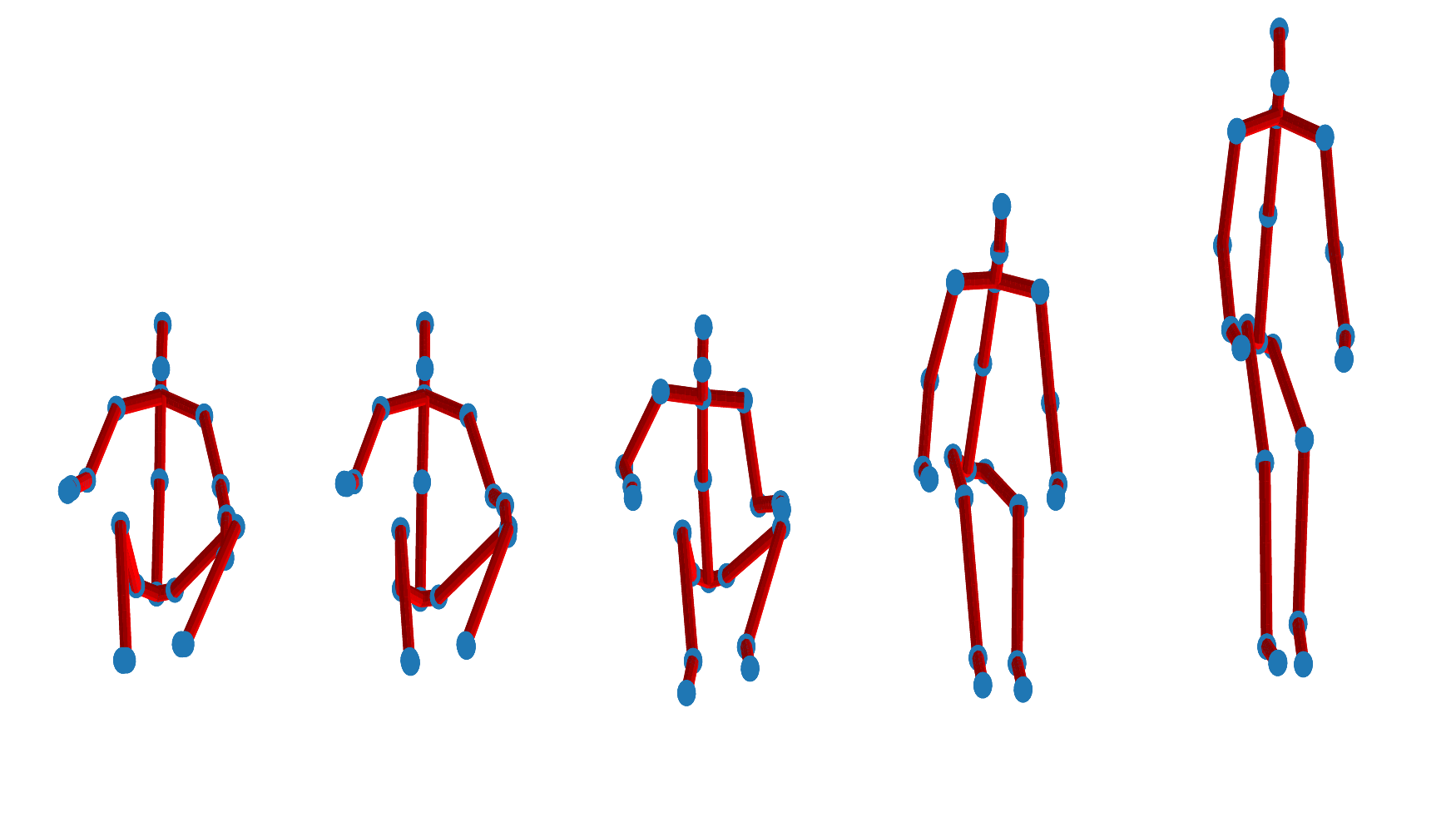}
        \caption*{Query: Stand Up}
        \label{}
    \end{subfigure}
\hfill
    \begin{subfigure}[b]{0.3\textwidth}
        \includegraphics[width=\textwidth]{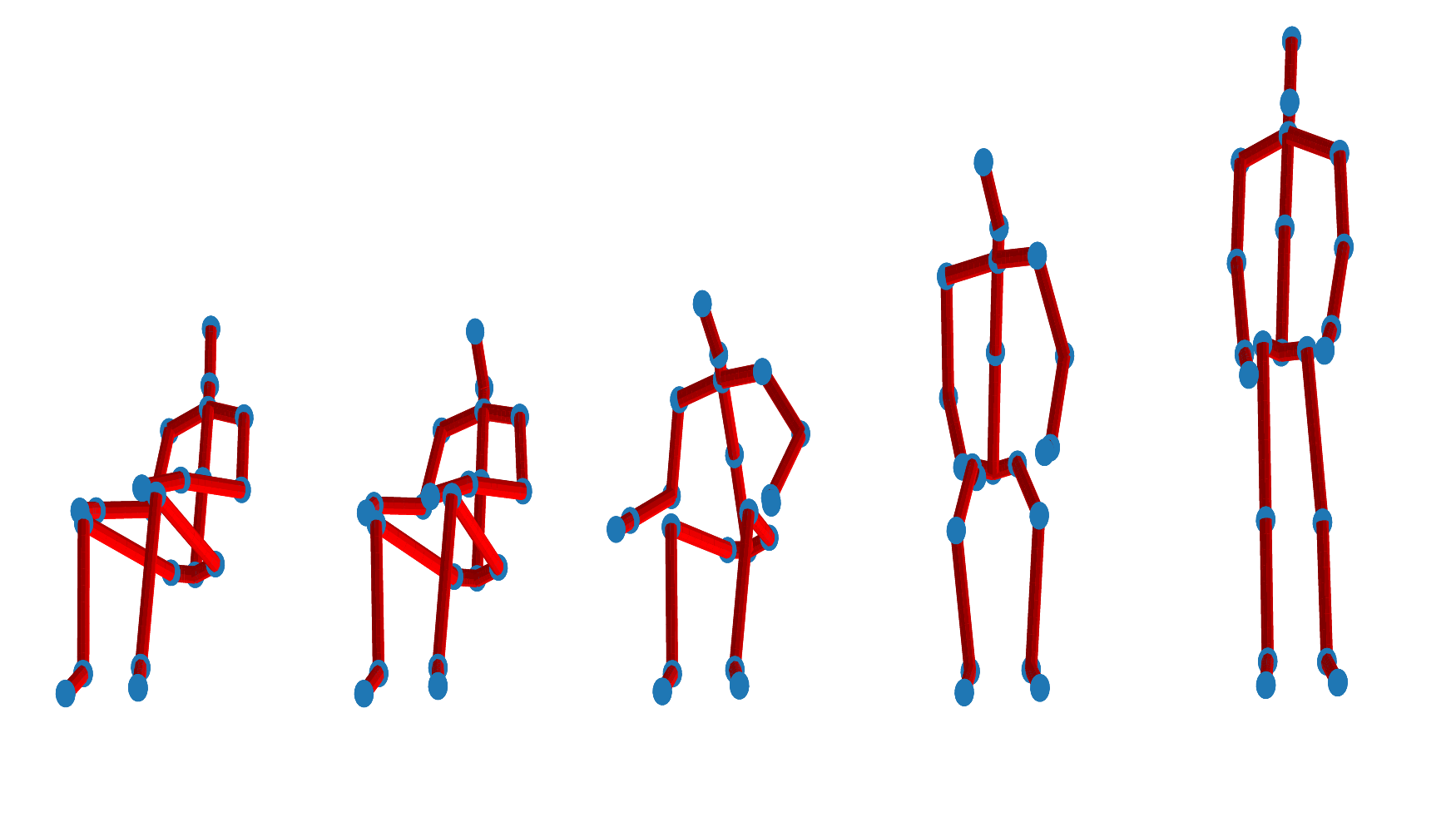}
        \caption*{Rank 1}
        \label{}
    \end{subfigure}
\hfill
    \begin{subfigure}[b]{0.3\textwidth}
        \includegraphics[width=\textwidth]{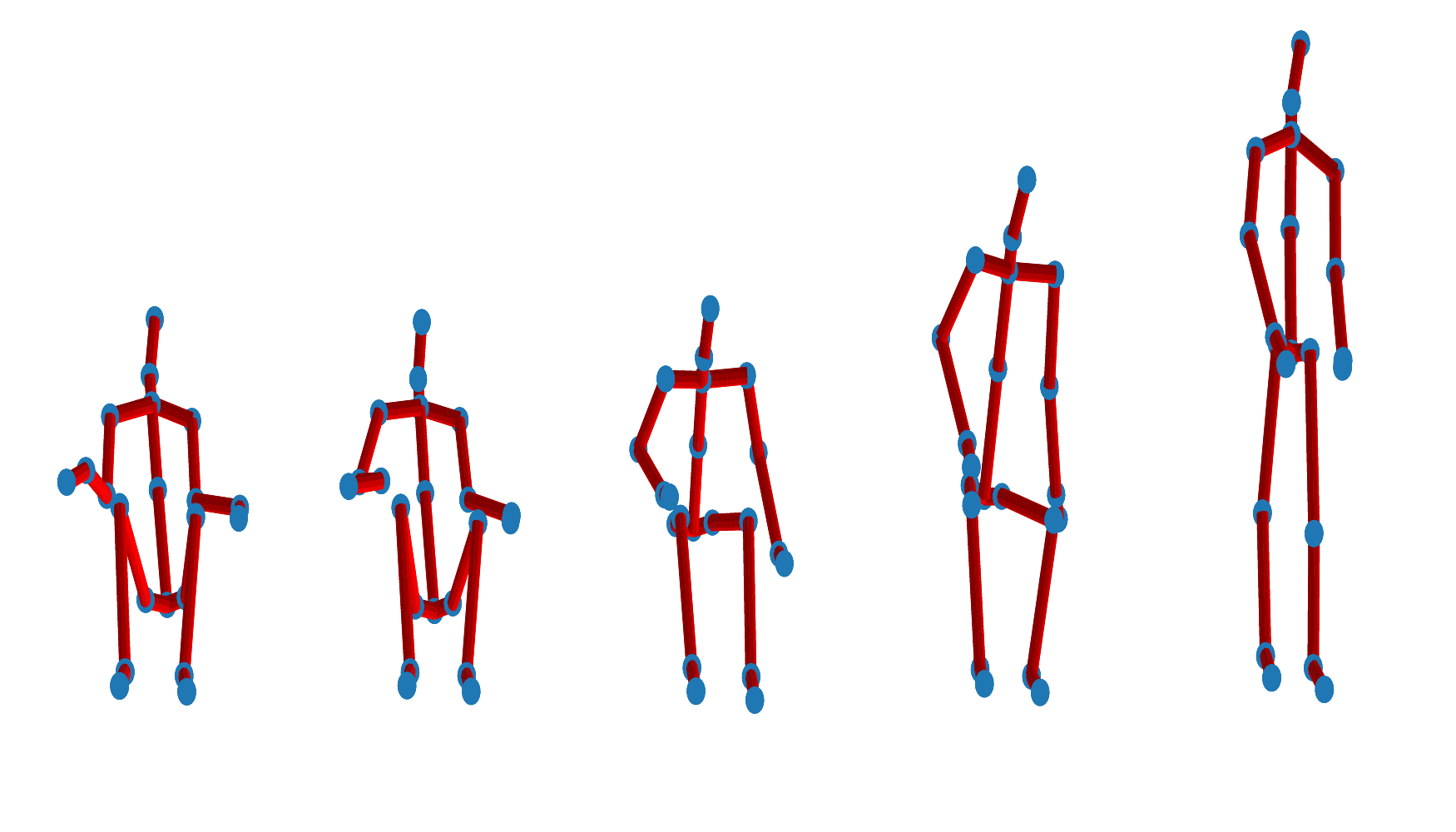}
        \caption*{Rank 2}
        \label{}
    \end{subfigure}

\end{center}
\caption{Retrieval results on NTU-RGB+D dataset \cite{shahroudy2016ntu} using our supervised setup.}
\label{fig:sup}
\end{figure*}

The results and inferences reported below are consistent for all datasets. For more detailed results, please see our supplementary video. Given a query, we show the top-2 ranked results in Figure~\ref{fig:self-sup}, \ref{fig:sup}, \ref{fig:noise} and \ref{fig:sub}.

\subsubsection{Results of Self-supervision.} Going beyond class labels, we are able to exploit inter-class information when trained with only self-supervision; therefore, the resulting retrieved motions are more closer to the query motion than the supervised setup, in terms of per frame error after \textit{DTW - 34mm of supervised v/s 31mm of unsupervised}. This is a promising result, particularly because existing datasets have a very limited number of labels and it enables us to exploit 3D sub-sequence similarity and perform retrieval in a label-invariant manner. 

\begin{center}
\begin{figure*}[h]
    \begin{subfigure}[h]{0.31\textwidth}
    \includegraphics[width=\linewidth]{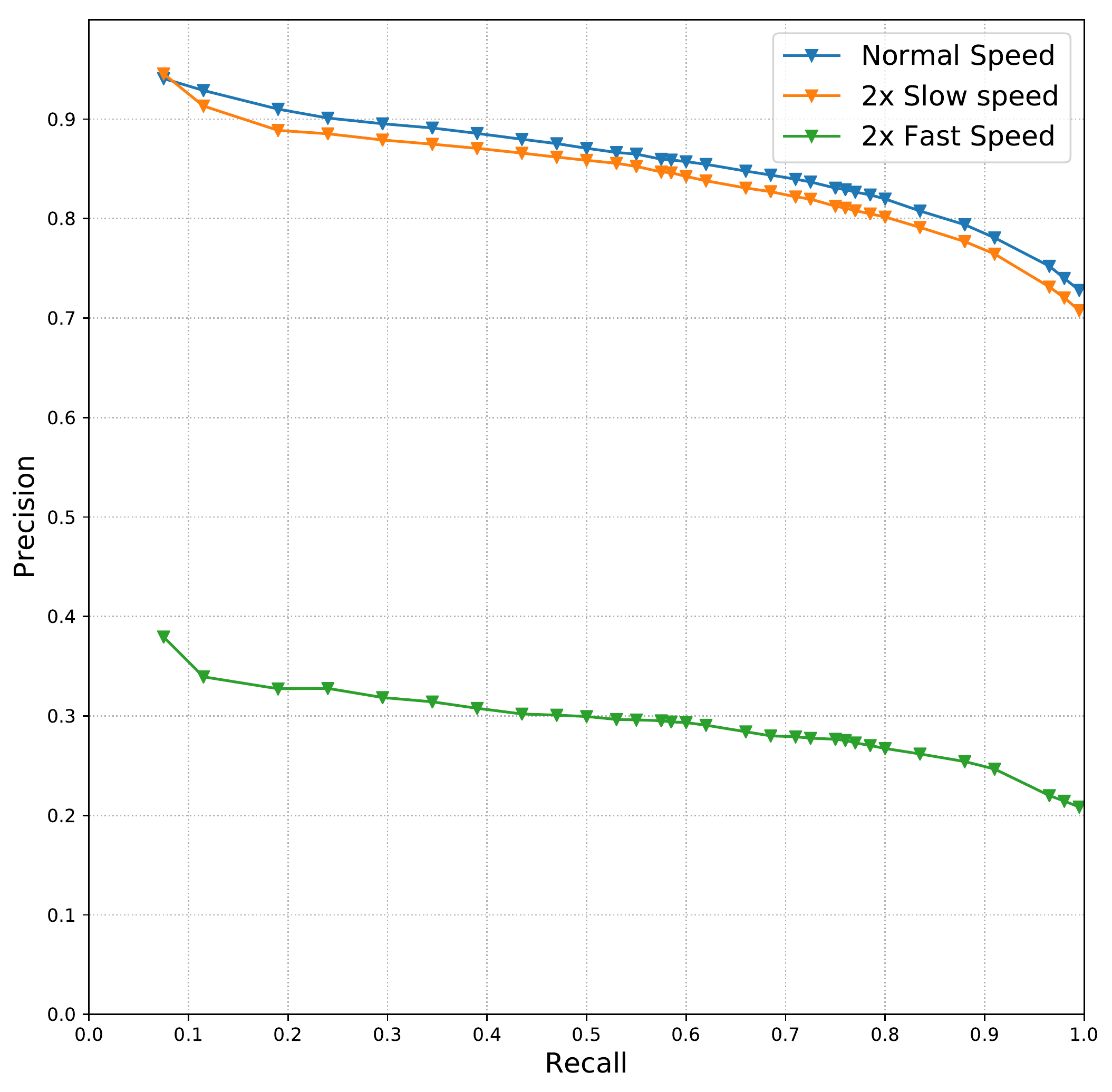}
    \caption{}
    \label{fig:pre_rec_sl_fa_no}
    \end{subfigure}
    \begin{subfigure}[h]{0.31\textwidth}
    \includegraphics[width=\linewidth]{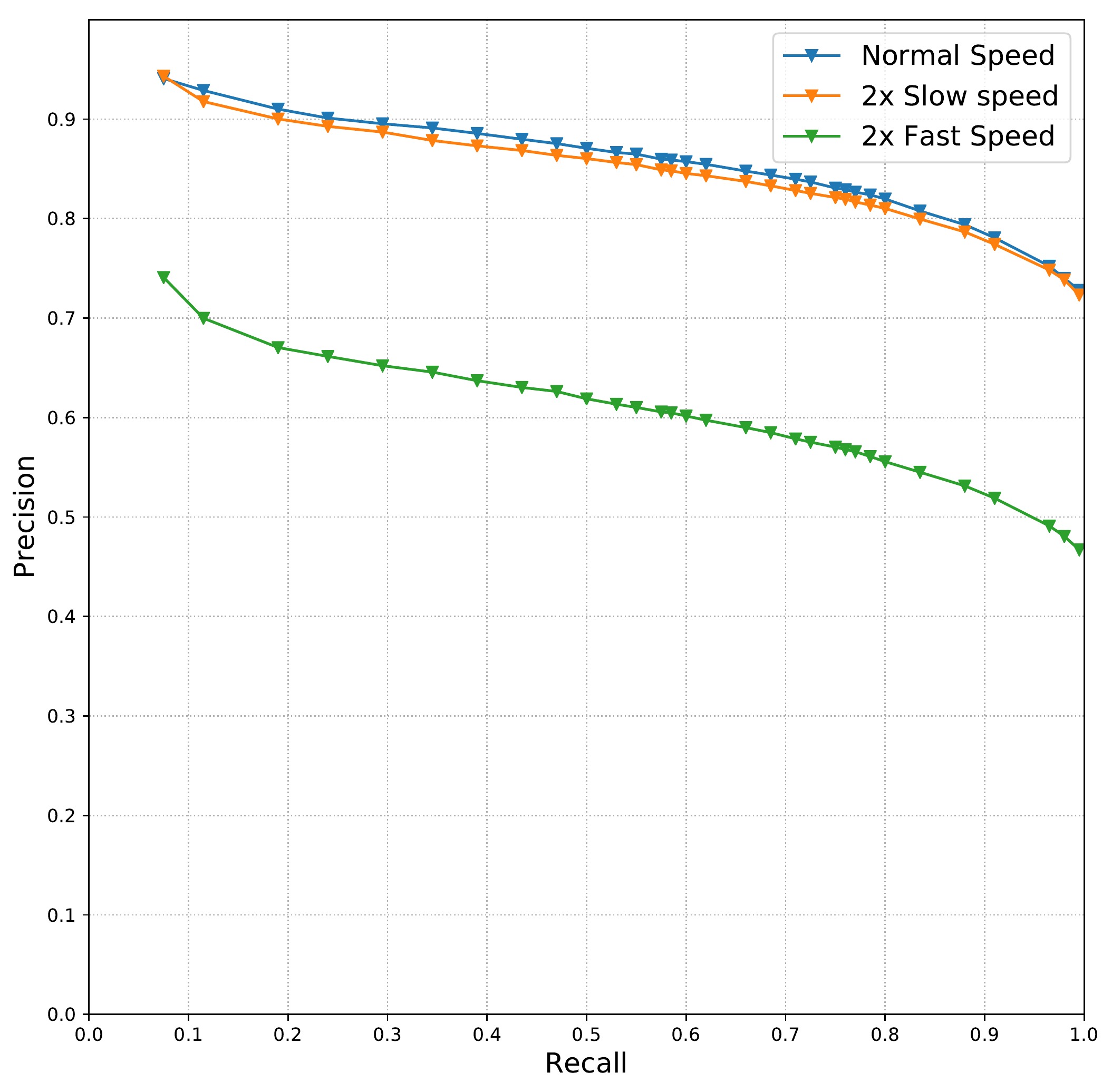}
    \caption{}
    \label{fig:pre_rec_sl_fa_tr}
    \end{subfigure}
    \begin{subfigure}[h]{0.31\textwidth}
    \includegraphics[width=\linewidth]{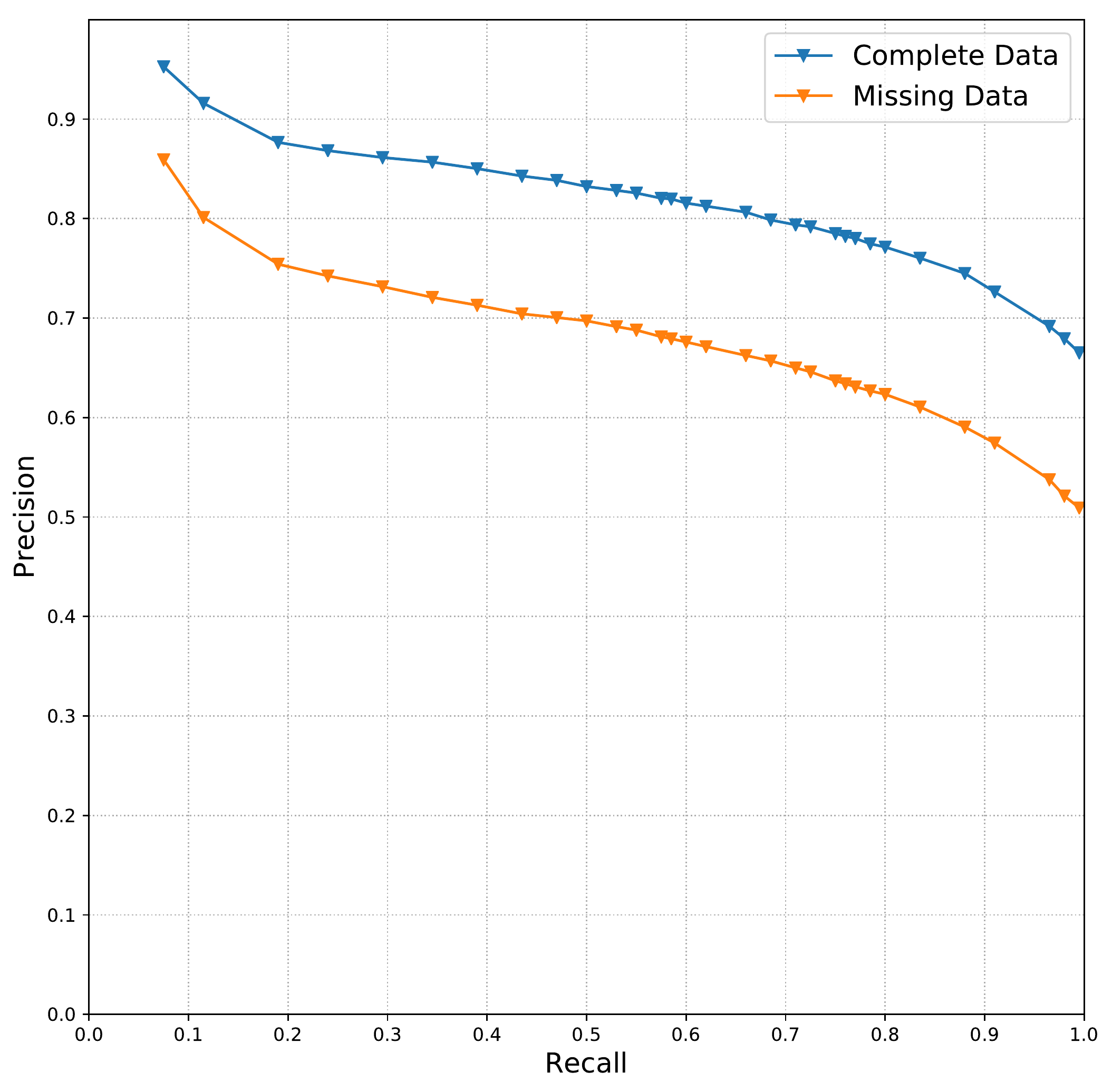}
    \caption{}
    \label{fig:pre_rec_noisy}
    \end{subfigure}
\caption{A comparison of our Precision Recall curves \textbf{(a)} before training for different speeds on HDM05 dataset, \textbf{(b)} after training for different speeds on HDM05 dataset and \textbf{(c)} with noisy data.}

\label{fig:short}
\end{figure*}
\end{center}

\begin{figure*}[h!]
\begin{center}
\hfil
    \begin{subfigure}[b]{0.3\textwidth}
        \includegraphics[width=\textwidth]{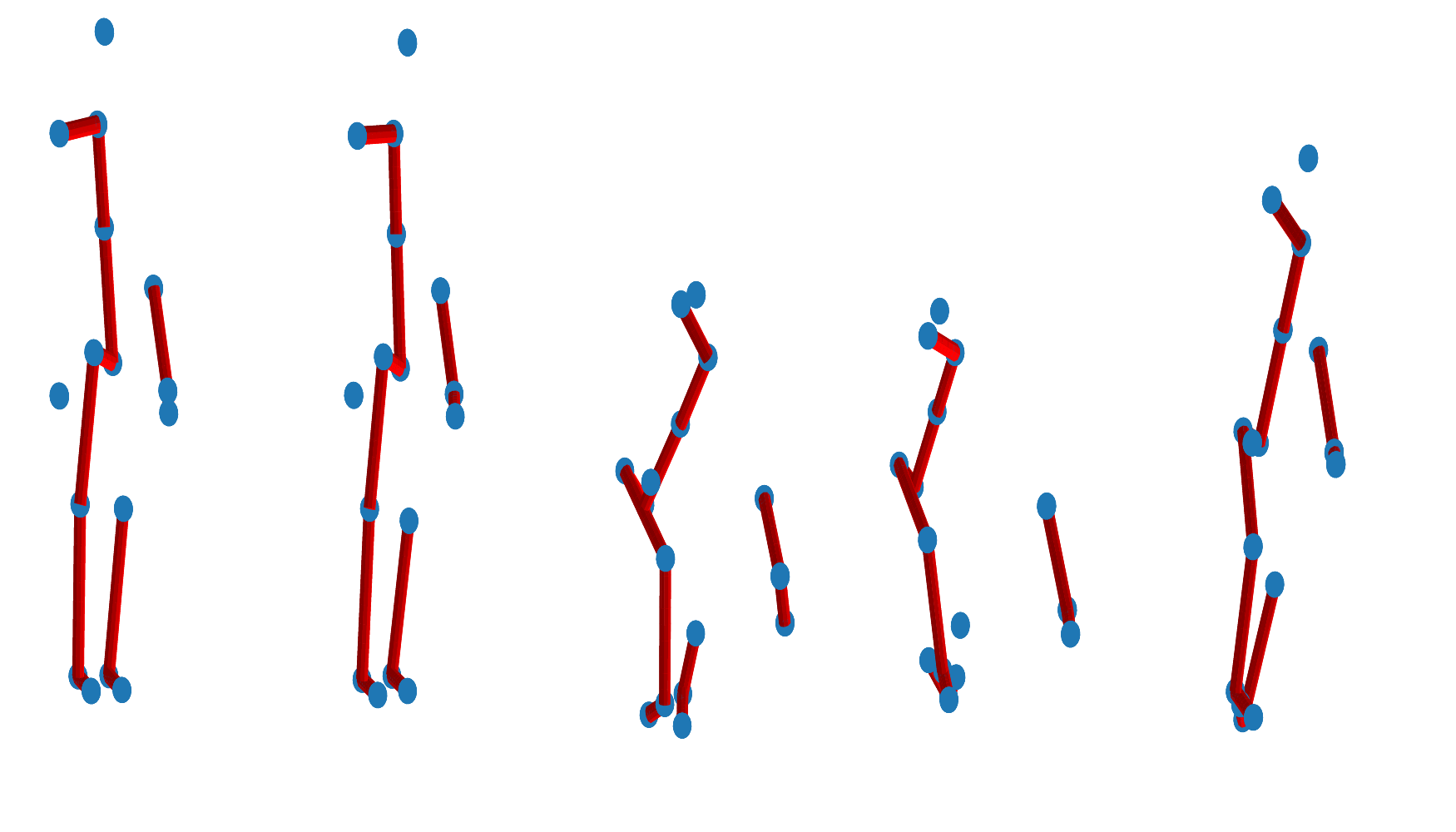}
        \caption*{Query: Pick Up}
        \label{}
    \end{subfigure}
\hfil
    \begin{subfigure}[b]{0.3\textwidth}
        \includegraphics[width=\textwidth]{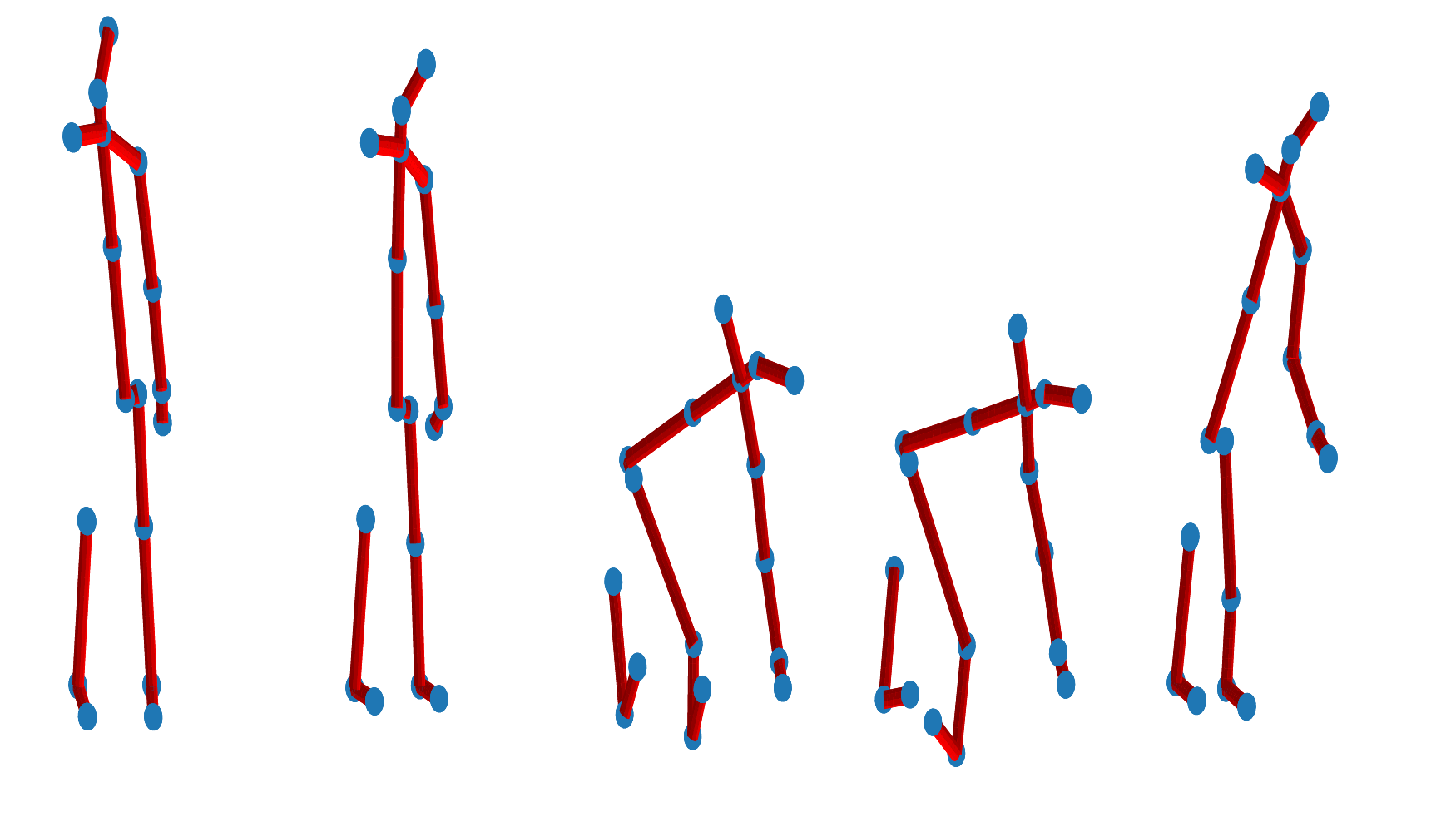}
        \caption*{Rank 1}
        \label{}
    \end{subfigure}
\hfil
    \begin{subfigure}[b]{0.3\textwidth}
        \includegraphics[width=\textwidth]{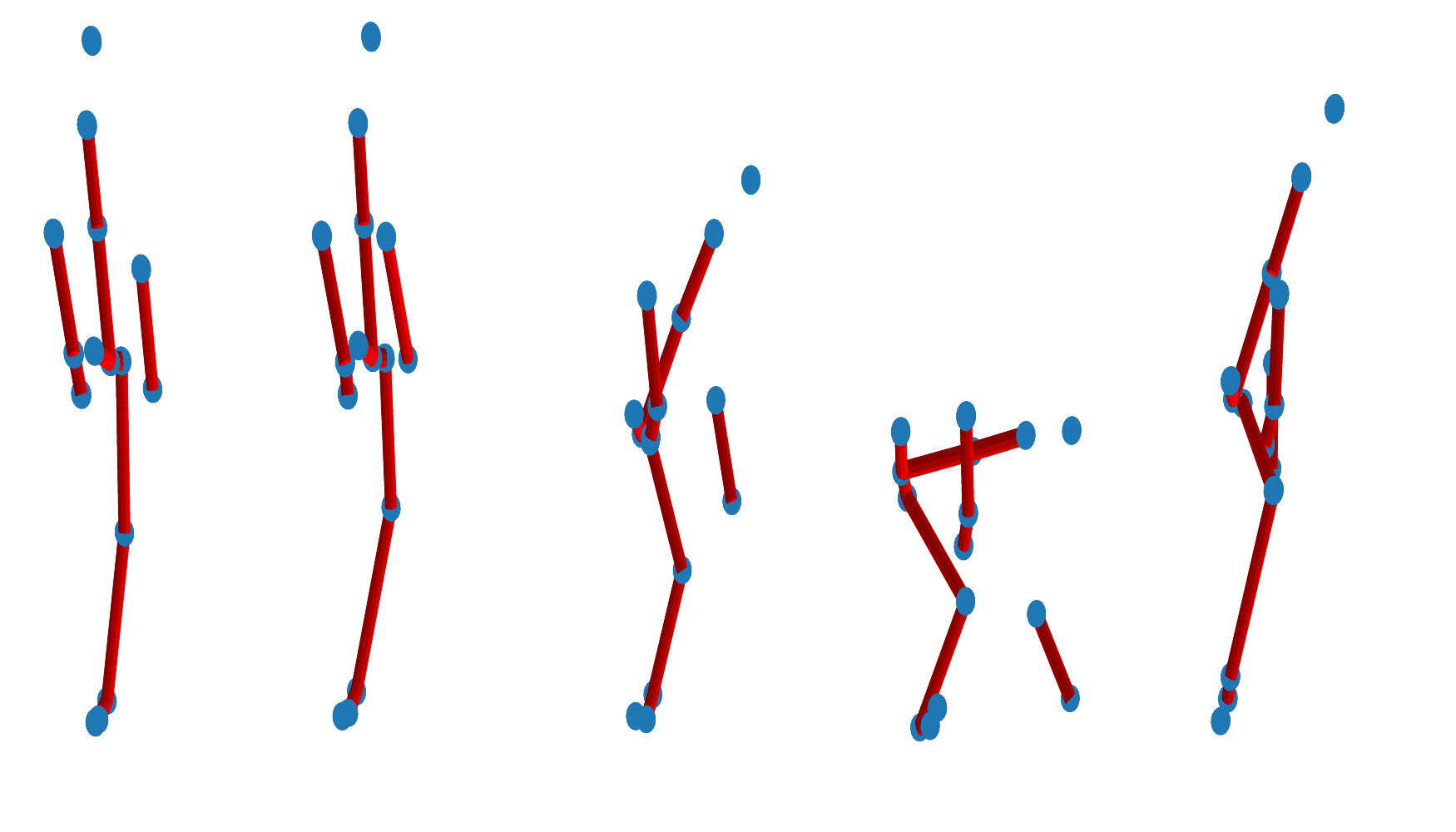}
        \caption*{Rank 2}
        \label{}
    \end{subfigure}
\hfil
\end{center}
\caption{Retrieval Results for Noisy Data}
\label{fig:noise}
\end{figure*}

\noindent \textbf{Sequences of different speeds. }Irrespective of the sampling rate/speed of motion, the motion field, and distance would be the same. So, we take care of motions performed at different speeds by minimizing all to the same embedding. We do this by simulating a sequence that is twice as slow and twice as short by interpolation and uniform sampling respectively and training a Siamese over them. Figure~\ref{fig:pre_rec_sl_fa_no} and~\ref{fig:pre_rec_sl_fa_tr} shows that more the number of frames, more amount of information is given to the network, and therefore, better the results. We handle short to very long sequences ranging in length from 15 to 600 frames. 


\newpage
\noindent \textbf{Noisy/Missing data. }To prove the robustness of our method towards noisy data, we trained and tested out model with missing data - random 20\% of joints missing from all frames of each sequence. This scenario simulates sensor noise or occlusions while detection of 3D skeletons. As shown in Figure~\ref{fig:pre_rec_noisy}, we still achieve an impressive retrieval accuracy in scenarios where optimization based state of the art methods would struggle.

\begin{figure*}[h!]
\begin{center}
    \begin{subfigure}[b]{0.8\textwidth}
        \includegraphics[width=\textwidth]{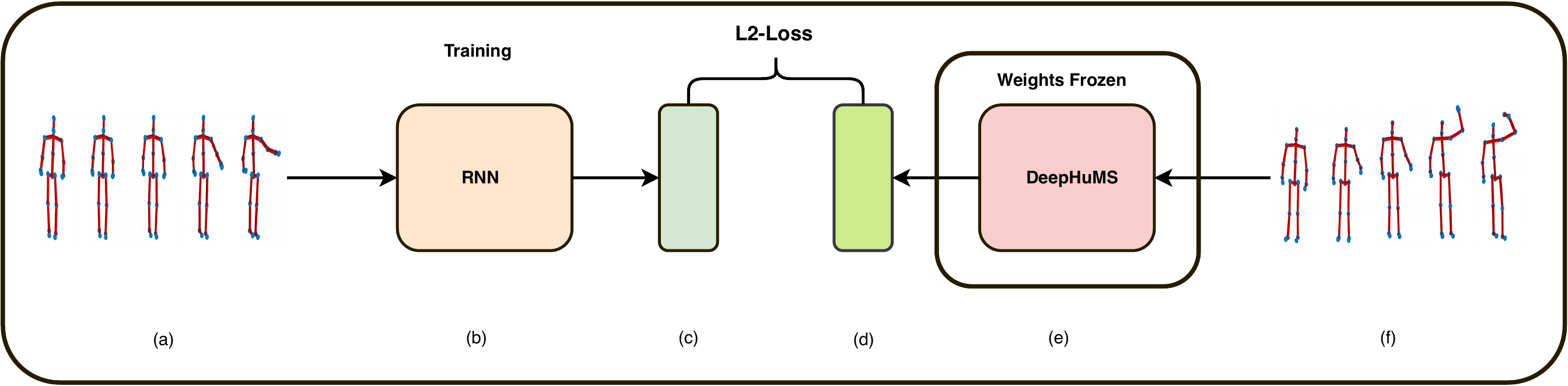}
        \label{fig:pre_rec_ra}
    \end{subfigure}

\caption{Given a subsequence (a) as the input, we use another RNN (b) to learn an embedding (c). This is minimized with L2 loss w.r.t the ground truth (d) generated from DeepHuMS (e) trained on long sequences (f)}
\label{fig:submodel}
\end{center}
\end{figure*}

\begin{figure*}[h!]
\begin{center}
    \begin{subfigure}[b]{0.3\textwidth}
        \includegraphics[width=\textwidth]{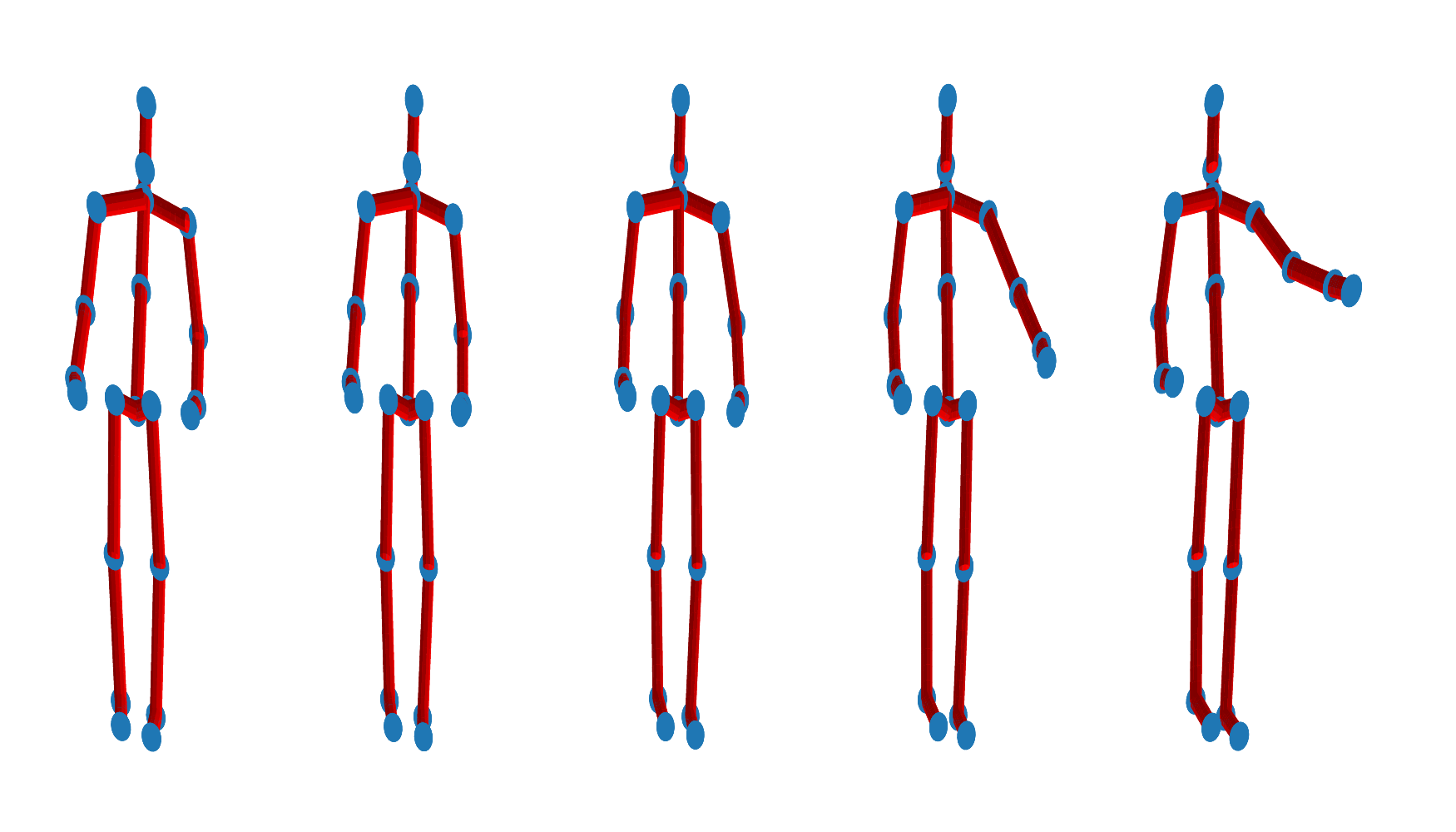}
        \caption*{Query: Raise Hand}
        \label{}
    \end{subfigure}
\hfill
    \begin{subfigure}[b]{0.3\textwidth}
        \includegraphics[width=\textwidth]{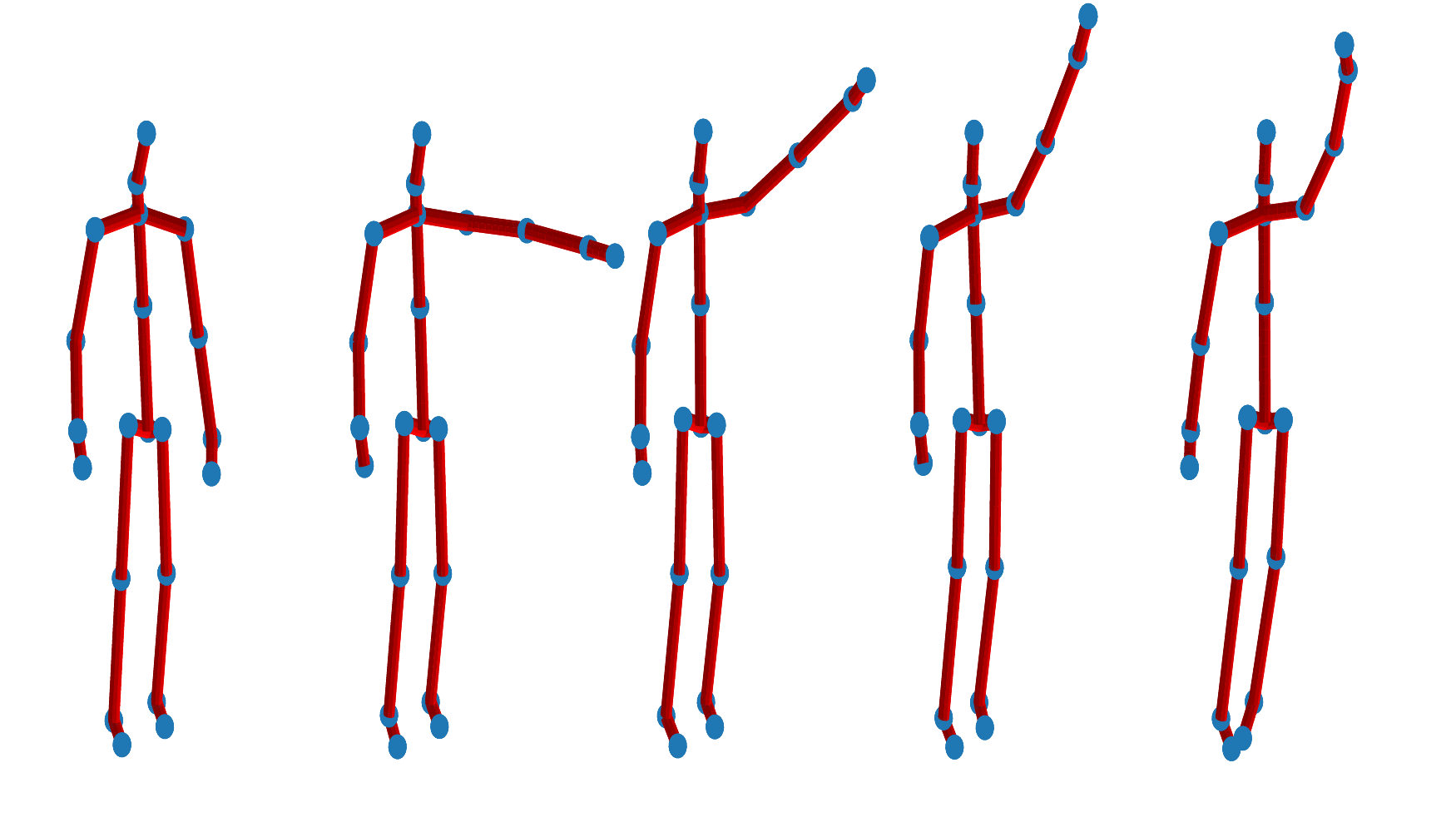}
        \caption*{Rank 1 (Wave)}
        \label{}
    \end{subfigure}
\hfill
    \begin{subfigure}[b]{0.3\textwidth}
        \includegraphics[width=\textwidth]{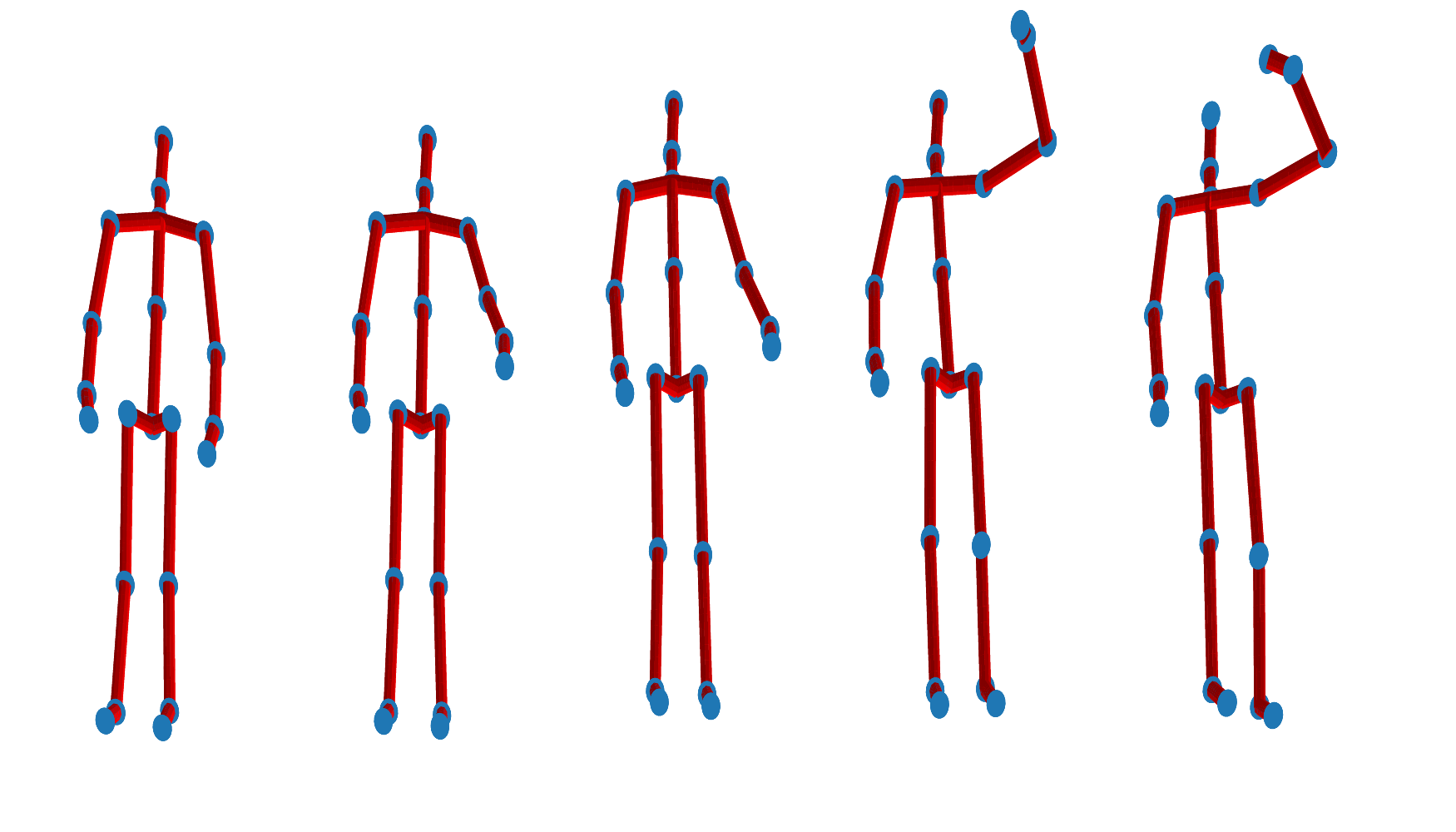}
        \caption*{Rank 2 (Salute)}
        \label{}
    \end{subfigure}

\end{center}
\caption{Retrieval results for sub sequence retrieval. Retrieved results are from different classes but query is a sub part of both of the retrieved videos.}
\label{fig:sub}
\end{figure*}

\noindent \textbf{Sub Motion Retrieval. }Sub Motion retrieval becomes important when we would like to search for a smaller action/motion in longer sequences. But it is a very challenging task due to the variations in length and actions in sub sequences. Moreover, our similarity metrics, in their current form can't account for sub sequences directly. To address this, we follow the model shown in Figure~\ref{fig:submodel}. Using this simple model, we retrieve the whole sequence it is a part of. This is a good starting point for the community and we believe that better solutions can be developed that directly incorporate sub-sequence information in the motion descriptor. 

\subsubsection{Retrieval Time. }We have a fairly low dimensional embedding of size 512, and perform a simple nearest neighbour search throughout the training dataset. This yields an average retrieval time for the test set to be 18ms for NTU RGB+D and 0.8ms for HDM05 dataset. The retrieval time is proportional to the dataset size, and one could use more advanced algorithms such as tree based searching. 

\subsection{Limitations}
Although our current model demonstrates impressive results, there exist some shortcomings in terms of generalisability and design. Firstly, indexing the sub-motion in the full sequence isn't trivial. Secondly, sequences with repetitive actions would be sub-optimal to handle with our full video-DeepHuMS descriptor. Both of these are because of different motion fields/distances as well as the lack of explicit temporal indexing of individual key-frames in the learned 3D human motion descriptor. In other words, we need either better utilization of the "semantic context" injected by the existing similarity metrics as well as need additional constructs to incorporate a better semantic context. This extends to a larger discussion about how to design models to learn in an unsupervised manner. 
\section{Conclusion}
In this paper, we make a case for using a learned representation for 3D Human Motion retrieval by means of a deep learning based model. Our model uses trajectory cues in a self-supervised manner to learn a generalizable, robust and discriminative descriptor. We overcome several of the limitations of current hand-crafted 4D motion descriptors such as their inability to handle  noisy/missing data, different speeds of the same motion, generalize to a large number of sequences and classes etc, thus making our model applicable to real world data. Lastly, we provide an initial model in the direction of 3D sub-motion retrieval, using the learned sequence descriptor as the ground truth. We compare with state-of-the-art 3D motion recognition as well as 3D motion retrieval methods on two large scale datasets - NTU RGB+D and HDM05 and demonstrate far superior performance on all fronts - class-wise retrieval accuracy, time and frame level distance.

\end{document}